\documentclass[10pt,twocolumn,letterpaper]{article}

\usepackage[pagenumbers]{cvpr} 

\usepackage{paralist}
\usepackage{pifont}
\usepackage{booktabs}
\usepackage{dsfont}
\usepackage{multirow}
\usepackage[most]{tcolorbox}
\usepackage{textcomp}
\newcommand{\xmark}{\ding{55}}  
\definecolor{cvprblue}{rgb}{0.21,0.49,0.74}
\usepackage[pagebackref,breaklinks,colorlinks,allcolors=cvprblue]{hyperref}

\title{Towards Explainable Bilingual Multimodal Misinformation Detection and Localization}

\author{
\parbox{\textwidth}{\centering
Yiwei He$^{1}$ \quad
Zhenglin Huang$^{1}$ \quad
Haiquan Wen$^{1}$ \quad
Tianxiao Li$^{1}$ \quad
Yi Dong$^{1}$ \\
Hao Fei$^{2}$ \quad
Baoyuan Wu$^{3}$ \quad
Guangliang Cheng$^{1\dagger}$\\[0.5em]
$^{1}$University of Liverpool, United Kingdom \quad
$^{2}$National University of Singapore, Singapore \\
$^{3}$The Chinese University of Hong Kong, Shenzhen, China \\[0.5em]
\textsuperscript{$\dagger$} Corresponding author. \quad
E-mail: \texttt{guangliang.cheng@liverpool.ac.uk} \\[0.5em]
}
}

\begin{document}
\maketitle
\begin{abstract}
The increasing realism of multimodal content has made misinformation more subtle and harder to detect, especially in news media where images are frequently paired with bilingual (e.g., Chinese-English) subtitles.  
Such content often includes localized image edits and cross-lingual inconsistencies that jointly distort meaning while remaining superficially plausible.  
We introduce \textbf{BiMi}, a bilingual multimodal framework that jointly performs region-level localization, cross-modal and cross-lingual consistency detection, and natural language explanation for misinformation analysis.  
To support generalization, BiMi integrates an online retrieval module that supplements model reasoning with up-to-date external context.  
We further release \textbf{BiMiBench}, a large-scale and comprehensive benchmark constructed by systematically editing real news images and subtitles, comprising \textbf{104,000} samples with realistic manipulations across visual and linguistic modalities.  
To enhance interpretability and cross-modal reasoning consistency, we apply Group Relative Policy Optimization (GRPO) to improve explanation quality, marking the first use of GRPO in this domain.  
Extensive experiments demonstrate that BiMi outperforms strong baselines by up to \textbf{+8.9} in classification accuracy, \textbf{+15.9} in localization accuracy, and \textbf{+2.5} in explanation BERTScore, advancing state-of-the-art performance in realistic, bilingual misinformation detection. Code, models, and datasets will be released.
\end{abstract}    
\section{Introduction}

\begin{figure*}[t]
\centering
\includegraphics[width=0.99\linewidth]{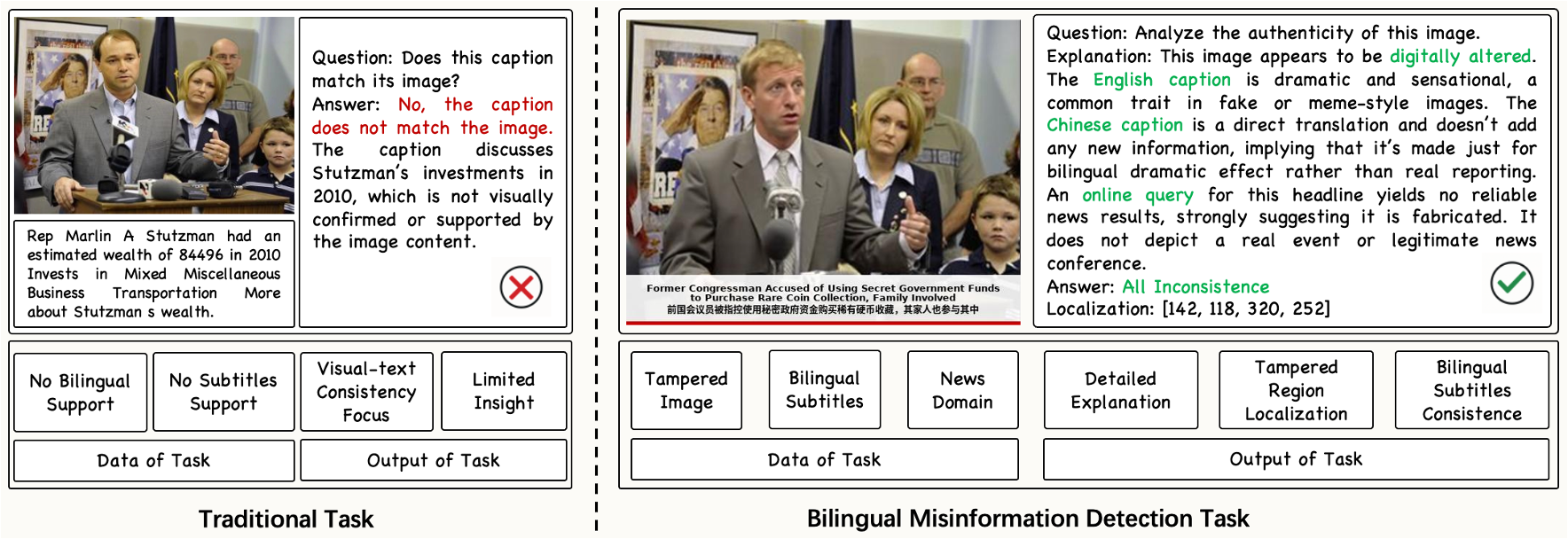}
\caption{Comparison of traditional vs. bilingual misinformation detection tasks. Traditional tasks focus on visual-text consistency with limited outputs (left). Our setting uses tampered images and bilingual subtitles, enabling richer outputs including region localization, cross-modal consistency, and explanation (right). \textit{Red indicates error, green indicates correctness. Best viewed in color.} }
\vspace{-4mm}
\label{fig:over}
\end{figure*}

The rapid progress of large generative models~\citep{achiam2023gpt,rombach2022high,zhao2025surveylargelanguagemodels,qwen,guo2025deepseek,hurst2024gpt} has substantially lowered the barrier to producing highly realistic multimodal content~\citep{openai2023dalle3}, enabling new creative applications but also heightening the risk of multimodal misinformation, where manipulated images and accompanying text jointly mislead audiences~\citep{qi2024sniffer,liu2024mmfakebench}.  A recent UNESCO report warns that the widespread adoption of generative-AI tools can be exploited to fabricate convincing visual–textual narratives, thereby creating fertile ground for large-scale multimodal misinformation.  Lost in Translation~\citep{quelle2025lost,guo2019future} further shows that misinformation frequently crosses language boundaries, underscoring the global risk of cross-lingual diffusion. A striking example occurred in early 2020, when a CDC report confirming the first U.S. case of COVID-19 community transmission was mistranslated on Chinese social media to claim that the virus originated in the United States, fueling public misunderstanding and geopolitical tension.  
These observations reveal how multimodal misinformation can exploit localized image edits and asymmetric subtitle translations to manipulate public perception across language communities~\citep{abdali2024multi}.  This creates an urgent need for methods capable of detecting fine-grained multimodal and cross-lingual inconsistencies—capabilities that remain largely underexplored. We therefore formulate the problem as bilingual multimodal misinformation detection: jointly localizing manipulated image regions and identifying cross-lingual inconsistencies while generating faithful natural-language explanations.

Despite notable progress in multimodal misinformation detection~\citep{wu2019misinformation,tahmasebi2024multimodal,wan2024dell} and deepfake detection~\cite{huang2025so,huang2025sida,zhao2025deepfakebench,shao2025deepfake}, existing approaches remain constrained by a predominant focus on coarse-grained image–text alignment and monolingual settings.  First, many systems match images and captions only at the global level~\citep{qi2024sniffer,liu2024mmfakebench}, leaving them unable to localize fine-grained, region-level manipulations such as localized edits or subtle subtitle tampering.  Second, their explanation modules typically produce high-level, generic rationales~\citep{shao2025deepfake}, providing little concrete evidence of why specific content is misleading.  Third, no prior work has specifically addressed bilingual subtitle inconsistencies, leaving the semantic divergence introduced by deliberately misleading cross-lingual translations largely unexplored. These limitations collectively underscore the need for a unified framework that can jointly reason over visual content and bilingual text with precise localization and faithful explanation—capabilities that motivate the approach we develop in this work.

However, building such a framework is technically demanding.
Achieving fine-grained grounding between image regions and textual cues remains beyond the capability of most existing multimodal large language models (MLLMs).
Moreover, producing faithful explanations in a bilingual multimodal setting requires accurate detection and articulation of inconsistencies across both modalities and languages.
Cross-lingual reasoning is further complicated by subtle semantic shifts that even fluent translations can introduce, often obscuring misinformation signals.
These challenges are especially acute in the news domain, where content evolves rapidly and exhibits high diversity and context dependence, making it non-trivial to design a system that jointly delivers precise localization, robust cross-lingual reasoning, and faithful explanation.

To address these challenges, we introduce BiMi, the first framework that jointly targets \textbf{bilingual subtitles inconsistency}, \textbf{region-level manipulation localization}, and \textbf{natural-language explanation generation}. Figure~\ref{fig:over} compares our unified task setup with prior multimodal approaches.
To improve generalization to emerging events, BiMi incorporates an online retrieval module that augments model reasoning with real-time external knowledge.
We further construct BiMiBench, the first large-scale benchmark for this setting, containing 104,000 news-image samples with realistic manipulations of visual content and bilingual subtitles.
Extensive experiments demonstrate that BiMi establishes a new state of the art, outperforming strong baselines by +8.9 in classification accuracy, +15.9 in localization accuracy, and +2.5 in explanation BERTScore, demonstrating robust overall improvements.

Our main contributions are as follows:
\begin{compactitem}
\item BiMiBench: the first large-scale benchmark for bilingual multimodal misinformation detection, comprising 104K news–image samples with fine-grained visual manipulations and bilingual subtitle inconsistencies.
\item BiMi Framework: a unified model that detects image–subtitle misinformation through region-level manipulation localization, multimodal and cross-lingual consistency detection, and natural-language explanation generation, directly addressing key limitations of prior work.
\item GRPO for Explanation: the first application of Group Relative Policy Optimization (GRPO)~\cite{shao2024deepseekmath} to improve the quality and faithfulness of bilingual multimodal explanations.
\item State-of-the-Art Results: BiMi achieves state-of-the-art performance on BiMiBench, with significant improvements in classification, localization, and explanation BERTScore over strong baselines.
\end{compactitem}

\section{Related Work}

\subsection{Misinformation Detection}
\begin{figure}[t!]
\centering
\includegraphics[width=0.9\linewidth]{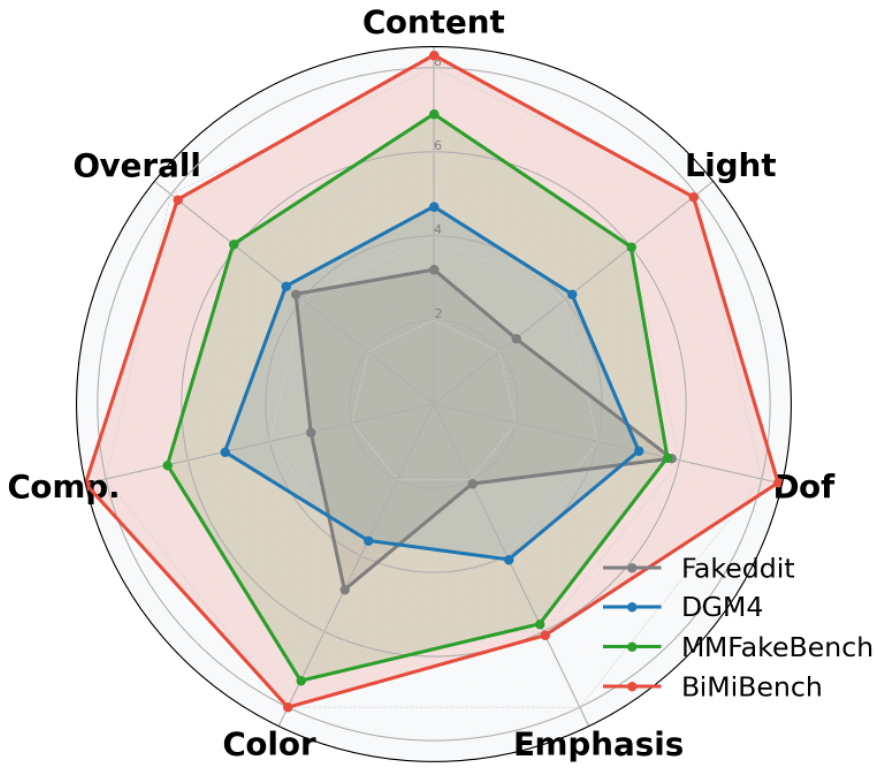}
\caption{Image quality comparison across datasets using the perceptual evaluation method of~\cite{chen2024evaluating}.
BiMiBench achieves higher visual quality across dimensions than prior datasets, indicating closer alignment with real-world social media imagery.}
\label{fig:bimibench_vis}
\vspace{-2mm}
\end{figure}

\textbf{Datasets.} Early misinformation detection datasets can be grouped into three main categories. Text-only: FEVER~\citep{Thorne18Fever} and FakeNewsNet~\citep{shu2020fakenewsnet} target textual fact verification and lack visual modality. Basic multimodal: Fakeddit~\cite{nakamura2019r}, MAIM~\citep{jaiswal2017multimedia}, and EMU~\citep{da-etal-2021-edited} pair images with text but are monolingual, use coarse edits, and provide no supervision for localization or explanation. Fine-grained multimodal: NewsCLIPpings~\citep{luo-etal-2021-newsclippings}, COSMOS~\citep{aneja2021cosmos}, MMFakeBench~\citep{liu2024mmfakebench}, and DGM4~\citep{shao2023detecting} enable richer reasoning yet still focus on caption-style inputs and ignore subtitle-level or cross-lingual inconsistencies. We introduce BiMiBench, which covers five types of visual and textual misinformation, uniquely supporting bilingual subtitles, region-level localization, and dual-modality tampering (Table~\ref{tab:dataset_half}); a perceptual analysis (Figure~\ref{fig:bimibench_vis}) shows it achieves higher realism and diversity than existing datasets.

\begin{table}[t!]
\centering
\scriptsize
\renewcommand{\arraystretch}{0.95}
\caption{Comparison of misinformation datasets. BiMiBench uniquely supports bilingual and localized manipulations.}
\vspace{-1mm}
\begin{tabular*}{\linewidth}{@{\extracolsep{\fill}} lccccccc @{}}
\toprule
\textbf{Dataset}  & \textbf{Sub.} & \textbf{Bi.} & \textbf{Text} &
\textbf{Vis.} & \textbf{Loc.} & \textbf{News} & \textbf{Mask} \\
\midrule
FEVER~\cite{Thorne18Fever}            & \xmark & \xmark & \checkmark & \xmark & \xmark & \xmark     & \xmark \\
FakeNewsNet~\cite{shu2020fakenewsnet} & \xmark & \xmark & \xmark     & \xmark & \xmark & \checkmark & \xmark \\
Fakeddit~\cite{nakamura2019r}         & \xmark & \xmark & \xmark     & \checkmark & \checkmark & \xmark & \xmark \\
MAIM~\cite{jaiswal2017multimedia}     & \xmark & \xmark & \checkmark & \xmark & \xmark & \xmark     & \xmark \\
EMU~\cite{da-etal-2021-edited}        & \xmark & \xmark & \xmark     & \checkmark & \checkmark & \xmark & \xmark \\
NewsCLIPpings~\cite{luo-etal-2021-newsclippings} & \xmark & \xmark & \xmark & \xmark & \xmark & \checkmark & \xmark \\
COSMOS~\cite{aneja2021cosmos}         & \xmark & \xmark & \xmark     & \xmark & \xmark & \xmark     & \xmark \\
DGM4~\cite{shao2023detecting}         & \xmark & \xmark & \checkmark & \checkmark & \checkmark & \checkmark & \xmark \\
MMFakeBench~\cite{liu2024mmfakebench} & \xmark & \xmark & \checkmark & \checkmark & \xmark & \xmark & \xmark \\
\midrule
\textbf{BiMiBench (Ours)} & \checkmark & \checkmark & \checkmark &
\checkmark & \checkmark & \checkmark & \checkmark \\
\bottomrule
\end{tabular*}
\vspace{-2mm}
\label{tab:dataset_half}
\end{table}

\noindent
\textbf{Methods.} Current multimodal misinformation methods mainly align image–text pairs for out-of-context detection~\cite{przybyla2020capturing,qi2021improving,shao2023detecting}.
NewsCLIPpings~\citep{luo-etal-2021-newsclippings} and COSMOS~\citep{aneja2021cosmos} use CLIP-based or contrastive learning but lack supervision for fine-grained localization.
SNIFFER~\citep{qi2024sniffer} and EMU~\cite{da-etal-2021-edited} add explainability via MLLMs yet rely on clean monolingual captions.
HAMMER~\citep{shao2023detecting} localizes manipulations but ignores subtitle-level or multilingual cues.
MMD-Agent~\citep{liu2024mmfakebench} and CroMe~\citep{choi2025cromemultimodalfakenews} broaden evaluation but remain limited to English captions.
In contrast, our framework is the first to handle tampered images with bilingual subtitles, supporting localization, reasoning and explanation.

\subsection{Multi-modal Large Language Models}

\textbf{Model.} Multimodal large language models (MLLMs) have greatly advanced cross-modal reasoning. Representative models such as Flamingo~\citep{alayrac2022flamingo}, BLIP-2~\citep{li2023blip}, MiniGPT-4~\citep{zhu2023minigpt4enhancingvisionlanguageunderstanding}, Gemini~\cite{gemini2023,comanici2025gemini}, InternVL-3~\citep{zhu2025internvl3exploringadvancedtraining}, Qwen-VL~\citep{bai2023qwenvlversatilevisionlanguagemodel,xu2025qwen2,qwen3}, Gpt-4o~\citep{hurst2024gpt} and LLaVA~\citep{liu2023llava,liu2024llavanext} extend pretrained LLMs with cross-modal attention and instruction tuning for vision-language tasks. DeepSeek-R1~\citep{guo2025deepseek,shao2024deepseekmath} further employs GRPO to enhance explanation quality. Yet most MLLMs focus on grounded understanding or generation~\citep{wu2025grounded,lin2025boosting} and are not trained to detect cross-modal inconsistencies. We adapt Gemma 3~\citep{team2025gemma} to strengthen reasoning on visual and bilingual cues for misinformation detection.

\noindent
\textbf{Training.} MLLMs training typically includes large-scale pre-training followed by post-training with supervised fine-tuning (SFT) and reinforcement learning from human feedback (RLHF)\citep{ouyang2022training}. RLHF aligns outputs with human preferences using a reward model; methods such as Proximal Policy Optimization (PPO)~\citep{schulman2017proximal}, Direct Preference Optimization (DPO)~\citep{rafailov2023direct}, and GRPO~\citep{guo2025deepseek,shao2024deepseekmath} refine policies through preference ranking. RLHF remains explored for detecting visual edits and bilingual subtitle inconsistencies.

\section{BiMiBench: A Benchmark for Bilingual Multimodal Misinformation}

Existing benchmarks for multimodal misinformation primarily focus on monolingual captions or synthetic mismatches, lacking realism, cross-lingual scope, and fine-grained supervision. In real-world scenarios, misinformation often involves tampered images with bilingual (Chinese-English) subtitles, where inconsistencies may occur in any modality. We introduce BiMiBench, a benchmark for multimodal misinformation detection with bilingual subtitles, providing labels and explanations for joint evaluation of classification, localization, and natural language explanation quality.

\subsection{Benchmark Construction}

\begin{figure*}[t!]
\centering
\includegraphics[width=0.99\linewidth]{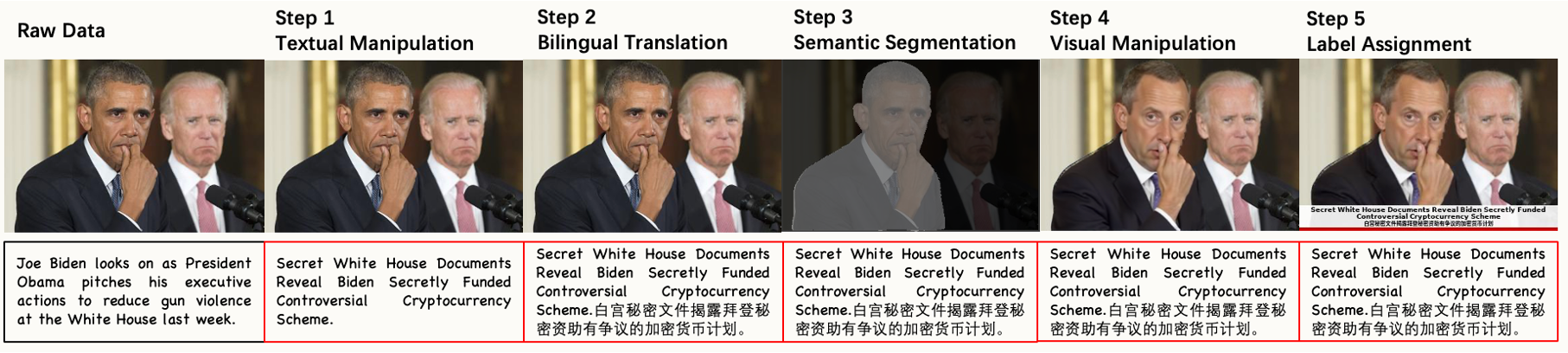}
\vspace{-1mm}
\caption{The data generation workflow used in constructing the BiMiBench benchmark.}
\vspace{-3mm}
\label{fig:overview}

\end{figure*}
\textbf{Details.} BiMiBench comprises 104,000 real-world news samples derived from the VisualNews~\citep{liu-etal-2021-visual} corpus, a professionally curated collection of image–text pairs with broad topical diversity and reliable editorial quality. Each sample pairs an image with Chinese–English subtitles; misinformation is introduced via localized image edits or Chinese/English subtitle modifications. About 80\% of the samples contain manipulations in at least one modality and 20\% serve as clean controls. Images (640×480–1024×768) and bilingual subtitles provide a challenging testbed for evaluating multimodal and bilingual reasoning.

\noindent
\textbf{Construction.} Our benchmark is constructed based on VisualNews, a large-scale image-text news dataset.

To generate realistic misinformation samples, we design a multi-step data generation pipeline shown in Figure~\ref{fig:overview}. Let \(\mathcal{D}_{\text{orig}} = \{(I_i, C_i)\}_{i=1}^N\) denote the original dataset, where \(I_i\) is an image and \(C_i\) is its associated English caption.

\textsc{\textit{Step 1: Textual Manipulation.}} Given an image-caption pair \((I_i, C_i)\) from VisualNews~\cite{liu-etal-2021-visual}, we use the Gemma 3~\cite{team2025gemma} model \(f_{\text{Gemma}}\) to generate a manipulated caption \(\tilde{C}_i = f_{\text{Gemma}}(I_i, C_i, \text{PROMPT})\), introducing inconsistencies while preserving contextual plausibility.

\textsc{\textit{Step 2. Bilingual Translation.}} We translate the original and manipulated English captions into Chinese using a translation API~\cite{googleapi2024} \(\mathcal{T}\), selected for its accuracy and reliability in large-scale bilingual news translation, yielding bilingual subtitle pairs: \(C^{\text{zh}}_i = \mathcal{T}(C_i), \tilde{C}^{\text{zh}}_i = \mathcal{T}(\tilde{C}_i)\).

\textsc{\textit{Step 3. Semantic Segmentation.}} To enable targeted visual editing, we use an instruction-based segmentation model LISA~\cite{lai2023lisa}:\(f_{\text{LISA}}\) to extract object masks based on the manipulated caption: \(M_i = f_{\text{LISA}}(I_i, \tilde{C}_i, \text{PROMPT})\), where \(M_i\) gets regions corresponding to entities in \(\tilde{C}_i\).

\textsc{\textit{Step 4. Visual Manipulation.}} We apply a visual manipulation function \(\mathcal{V}\) to perform localized editing on the original image using the object masks: \(\tilde{I}_i = \mathcal{V}(I_i, M_i)\), where the edited image \(\tilde{I}_i\) semantically aligns with the manipulated caption \(\tilde{C}_i\). To ensure visual diversity and realism, we adopt a mix of recent state-of-the-art image editing techniques, including FLUX~\cite{flux2024}, VAR~\cite{tian2024visual}, and SDXL~\cite{podell2023sdxlimprovinglatentdiffusion}.

\textsc{\textit{Step 5. Label Assignment.}} For each final sample, we randomly select real or manipulated versions of the image and subtitles to construct a multimodal example: \( S_i^*=S\left(I_i^*,C_i^{\text{en}},L_i^{\text{zh}}\right), I_i^* \in \{I_i, \tilde{I}_i\}, C_i^{\text{en}} \in \{C_i, \tilde{C}_i\}, L_i^{\text{zh}} \in \{C^{\text{zh}}_i, \tilde{C}^{\text{zh}}_i\}\), where misinformation may appear in any modality individually or in combination.

\begin{table}[h]
\centering
\small
\renewcommand{\arraystretch}{1.0}
\caption{BiMiBench Category Definitions.}
\begin{tabular*}{\linewidth}{@{\extracolsep{\fill}} l l @{}}
\toprule
\textbf{Category} & \textbf{Definition} \\
\midrule
All Consistent & \(I_i^* = I_i,\, C_i^{\mathrm{en}} = C_i,\, L_i^{\mathrm{zh}} = C_i^{\mathrm{zh}}\) \\
Image Manipulated & \(I_i^* = \tilde{I}_i,\, C_i^{\mathrm{en}} \ne \tilde{C}_i\) or \(L_i^{\mathrm{zh}} \ne \tilde{C}_i^{\mathrm{zh}}\) \\
Both Misaligned & \(I_i^* = I_i,\, C_i^{\mathrm{en}} = \tilde{C}_i,\, L_i^{\mathrm{zh}} = \tilde{C}_i^{\mathrm{zh}}\) \\
Chinese Misaligned & \(I_i^* = I_i,\, C_i^{\mathrm{en}} = C_i,\, L_i^{\mathrm{zh}} = \tilde{C}_i^{\mathrm{zh}}\) \\
English Misaligned & \(I_i^* = I_i,\, C_i^{\mathrm{en}} = \tilde{C}_i,\, L_i^{\mathrm{zh}} = C_i^{\mathrm{zh}}\) \\
All Inconsistent & \(I_i^* = \tilde{I}_i,\, C_i^{\mathrm{en}} = \tilde{C}_i,\, L_i^{\mathrm{zh}} = \tilde{C}_i^{\mathrm{zh}}\) \\
\bottomrule
\end{tabular*}
\label{tab:category_inline}
\end{table}

\textbf{Categories.} Each BiMiBench sample is labeled into one of six categories according to consistency among image \(I_i^*\), English subtitle \(C_i^{\text{en}}\), and Chinese subtitle \(L_i^{\text{zh}}\). (1) \textit{All Consistent}: all modalities agree; (2) \textit{Image Manipulated}: image tampered, at least one subtitle mismatched; (3) \textit{Both Misaligned}: image real, both subtitles misleading; (4) \textit{Chinese Misaligned}: only Chinese subtitle misleading; (5) \textit{English Misaligned}: only English subtitle misleading; (6) \textit{All Inconsistent}: image tampered and both subtitles misleading. Formal definitions show in Table~\ref{tab:category_inline}.

All samples were manually reviewed to ensure annotation quality and consistency. 
Each item was independently checked by two annotators with a senior reviewer resolving disagreements, following predefined guidelines on factual correctness and translation fidelity.
To address potential ethical and privacy concerns, we used only publicly available news content, removed any personally identifiable information, and release the dataset solely for non-commercial research in accordance with the source licenses.
This design enables structured supervision of complex multimodal misinformation scenarios while maintaining high standards of data integrity and ethical compliance. Detailed information about the BiMiBench can be found in Appendix.
\section{BiMi: A Bilingual Multimodal Misinformation Detection Framework}

Real-world multimodal misinformation often involves image edits and inconsistencies between bilingual subtitles. Existing models struggle to detect such cross-modal manipulations and lack adaptability to emerging events. We propose BiMi, a bilingual multimodal framework that localizes manipulated regions, detects cross-modal inconsistencies, and generates explanations. To enhance generalization, BiMi integrates an online retrieval module that provides real-time external context. 

\begin{figure*}[t!]
\centering
\vspace{-1mm}
\includegraphics[width=0.99\linewidth]{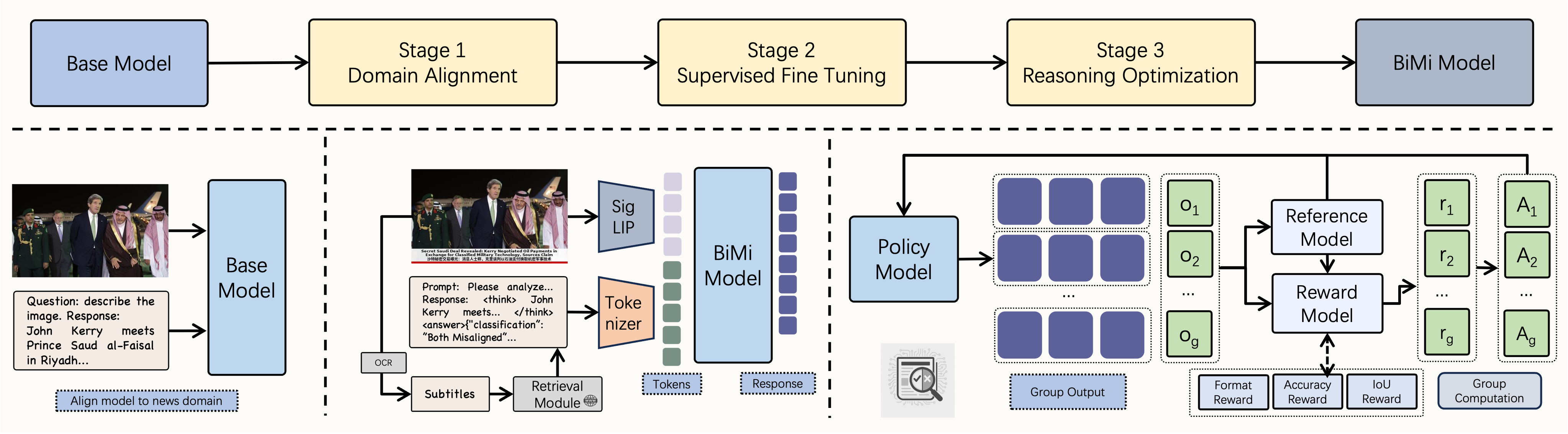}
\caption{The overview of the training strategy. Three stages: domain alignment on news data, instruction tuning with task-specific prompts, and GRPO optimization with structured rewards.}
\vspace{-4mm}
\label{fig:method}

\end{figure*}

\subsection{Framework}

\textbf{Modeling.} We adopt Gemma 3~\cite{team2025gemma} as BiMi’s backbone for its strong multilingual understanding and vision–language alignment, enabling detection of subtle manipulations across images and bilingual subtitles. The input image with overlaid Chinese and English subtitles is encoded into patch-level embeddings, which are fused through attention layers to create a unified multimodal context. This design allows joint reasoning over visual and bilingual textual cues to identify object manipulation, semantic shifts, and cross-lingual inconsistencies.

\noindent
\textbf{Retrieval Module.} To enhance adaptability to emerging misinformation, BiMi employs a retrieval module at inference. Chinese and English subtitles are extracted via OCR to form a bilingual query to the Google Search API~\citep{googleapi2024}; the top-3 retrieved passages are prepended to the input as a unified prompt $\mathcal{P}=\operatorname{concat}(R,I,S)$, where $R$, $I$, and $S$ denote the retrieved context, image, and subtitles. This auxiliary context supplies timely external information that improves the model’s generalization to previously unseen or rapidly evolving misinformation, while the final predictions remain grounded in the image–subtitle content.

\subsection{Multi-stage Training}

To effectively adapt the pretrained Gemma to the task of multilingual multimodal misinformation detection, we adopt a three-stage training strategy aimed at progressively aligning the model with the news domain, task-specific instructions, and high-quality reasoning objectives. Figure~\ref{fig:method} shows the overview of the training strategy.

\noindent
\textbf{Stage 1: Domain Alignment.}  
To adapt the model to the linguistic and visual traits of the news domain, we perform instruction-based tuning on VisualNews~\cite{liu-etal-2021-visual} image–caption pairs. Each instance uses an instruction prompt (e.g., \texttt{Describe this image in English/Chinese}) with an English or translated Chinese caption. This trains the model to generate captions from images and prompts, strengthening visual–text grounding and bilingual reasoning for downstream misinformation detection.

\noindent
\textbf{Stage 2: Instruction Tuning.}  
We fine-tune the model with single-turn instruction-following data tailored for multimodal misinformation detection. Each sample includes an image \(X_v\), a bilingual subtitle pair, and a prompt \(X_{\text{q}}\) that asks the model to perform multiple sub-tasks: detect manipulation, assess cross-modal consistency, and explain the decision. The assistant response \(X_{\text{a}}\) includes structured answers and a natural language explanation. We follow a unified sequence format:
\[
X = \texttt{<system>} \, [X_{\text{q}}; X_v] \, \texttt{<STOP>} \, X_{\text{a}} \, \texttt{<STOP>}
\]
where \(X_{\text{q}}\) contains the natural language instruction and bilingual subtitles, and \(X_{\text{a}}\) consists of three binary labels and a free-form explanation, formatted as:
\[
  \texttt{<think>} \, E \, \texttt{</think>} \quad \texttt{<answer>} \, y \, \texttt{</answer>}
\]
The model is trained to generate \(X_{\text{a}}\) conditioned on \(X_{\text{q}}\) and \(X_v\), by maximizing the likelihood:
\begin{equation}
P(X_{\text{a}} \mid X_{\text{q}}, X_v) = \prod_{i=1}^{L} P_\theta(x_i \mid x_{<i})
\end{equation}
where \(x_i\) denotes the \(i\)-th token in the assistant response. Only the tokens in \(X_{\text{a}}\) are used to compute the loss. This tuning step encourages the model to align with the structure and reasoning required for misinformation detection.

\noindent
\textbf{Stage 3: GRPO-based Reasoning Optimization.}
To enhance reasoning and explanation quality for multimodal misinformation detection, we adopt Group Relative Policy Optimization (GRPO)~\cite{shao2024deepseekmath}. GRPO ranks candidate outputs within each batch, which fits our setting where explanations for subtle visual edits and bilingual subtitle inconsistencies may be partially correct, and avoids the reward-model sensitivity of PPO and the pairwise preference assumption of DPO.

Given a question \(q\), GRPO samples \(N\) candidate responses \(\{r_1, r_2, \ldots, r_N\}\) from the policy \(\pi_\theta\) and evaluates each response \(o_i\) using a reward function \(R\left(q, o_i \right)\), which measures the quality of the candidate in the context of the given question. GRPO encourages the model to generate responses with higher advantages within the group by updating the policy \(\pi_\theta\) using the following objective:

\begin{equation}
\begin{aligned}
\mathcal{J}_{\mathrm{GRPO}}(\theta)
&=
\mathbb{E}_{\{ o_i \}_{i=1}^N \sim \pi_{\theta_{\text{old}}}(q)}
\bigg[
\frac{1}{N} \sum_{i=1}^N
\Big(
\min\!\big( s_1 A_i,\; s_2 A_i \big)
\\
&\qquad
- \beta\, \mathbb{D}_{\mathrm{KL}}\!\big( \pi_\theta \,\|\, \pi_{\mathrm{ref}} \big)
\Big)
\bigg].
\end{aligned}
\label{eq:grpo-objective}
\end{equation}
\begin{equation}
\begin{aligned}
A_i &= 
\frac{
r_i - \mathrm{mean}(r_1, \ldots, r_N)
}{
\mathrm{std}(r_1, \ldots, r_N)
}, \\
s_1 &= 
\frac{
\pi_\theta(o_i \mid q)
}{
\pi_{\theta_{\text{old}}}(o_i \mid q)
}, \qquad
s_2 = \mathrm{clip}\!\left( s_1,\; 1 - \epsilon,\; 1 + \epsilon \right).
\end{aligned}
\label{eq:advantage-weights}
\end{equation}

\noindent Where \(A_i\) represents the advantage of the candidate response \(o_i\) relative to other sampled responses.
Following DeepSeek-R1, we use both format and accuracy reward.

\noindent
\textbf{Reward function.} To optimize the reasoning ability and explanation quality of BiMi during the final training stage, we design a composite reward function tailored for GRPO. In our setting, we focus on three core tasks: misinformation classification, tampered region localization, and natural language explanation generation. 


\textit{Format reward.} To enforce structured outputs, we define a format reward \(R_{\text{format}}\) that equals 1 when the model output follows the predefined format with \texttt{<answer>}\texttt{</answer>} and optional \texttt{<think>}\texttt{</think>} tags, and 0 otherwise: $R_{\text{format}}=\mathds{1}\!\left[\text{output matches expected format}\right].$
\textit{Localization reward.} We define an IoU-based reward \(R_{\text{Loc}} = \frac{|\mathcal{M}_{\text{pred}} \cap \mathcal{M}_{\text{gt}}|}{|\mathcal{M}_{\text{pred}} \cup \mathcal{M}_{\text{gt}}|}\), where \(\mathcal{M}_{\text{pred}}\) and \(\mathcal{M}_{\text{gt}}\) are the predicted and ground-truth bounding boxes of the tampered region.
\textit{Classification reward.} We define $R_{\text{cls}} = \mathds{1}\!\left[C_{\text{pred}} = C_{\text{gt}}\right].$ to give a reward of 1 when the predicted label matches the ground truth and 0 otherwise.

The final reward combines all task-specific objectives, including formatting, classification, localization: \(R_{\text{total}} = R_{\text{format}} + R_{\text{cls}} + R_{\text{loc}}\). This unified reward encourages the model to generate structured, accurate, and interpretable predictions across modalities without introducing any additional weighting coefficients, so each objective contributes equally.

This progressive training strategy equips BiMi with the capability to perform fine-grained classification, localization, and explanation, and is designed to support generalization to challenging, real-world misinformation cases.

\section{Experiments}

\subsection{Experimental Setup}

\begin{table*}[t]
\centering
\small
\setlength{\tabcolsep}{3.8pt}
\renewcommand{\arraystretch}{1.1}
\caption{Combined results on BiMiBench and MMFakeBench (metrics in \%). IoU is reported when region-level localization is available (“–” denotes not applicable). All results are averaged over three runs with standard deviation below 0.5\%. \textit{Bold and underline indicate best and second-best performance, respectively.}}

\begin{tabular}{llcccccc}
\toprule
\multirow{2}{*}{Method} & \multirow{2}{*}{MLLM} &
\multicolumn{4}{c}{BiMiBench} &
\multicolumn{2}{c}{MMFakeBench} \\
\cmidrule(lr){3-6} \cmidrule(lr){7-8}
& & ACC & F1 & IoU & BERT & ACC & F1 \\
\midrule

InternVL3~\cite{zhu2025internvl3exploringadvancedtraining} & Qwen2.5-7B
& 20.85 {\color{red}(-25.95)}
& 18.42 {\color{red}(-24.37)}
& 7.23  {\color{red}(-23.01)}
& 71.82 {\color{red}(-11.08)}
& 42.69 {\color{red}(-28.25)}
& 34.80 {\color{red}(-27.69)} \\

InternVL3~\cite{zhu2025internvl3exploringadvancedtraining} & Qwen2.5-14B
& 28.47 {\color{red}(-18.33)}
& 26.71 {\color{red}(-16.08)}
& \underline{14.38} {\color{red}(-15.86)}
& 76.54 {\color{red}(-6.36)}
& -- & -- \\

Qwen3~\cite{qwen3} & Qwen3-8B
& 21.43 {\color{red}(-25.37)}
& 14.96 {\color{red}(-27.83)}
& 6.28 {\color{red}(-23.96)}
& 73.92 {\color{red}(-8.98)}
& 44.82 {\color{red}(-26.12)}
& 39.49 {\color{red}(-23.00)} \\

LLaVA-1.6~\cite{liu2024llavanext} & Vicuna-7B
& 22.91 {\color{red}(-23.89)}
& 20.74 {\color{red}(-22.05)}
& 8.91 {\color{red}(-21.33)}
& 75.77 {\color{red}(-7.13)}
& 30.09 {\color{red}(-40.85)}
& 24.31 {\color{red}(-38.18)} \\

Llama 3.1~\cite{grattafiori2024llama3herdmodels} & Llama-3.1-8B
& 24.61 {\color{red}(-22.19)}
& 21.94 {\color{red}(-20.85)}
& 5.27 {\color{red}(-25.0)}
& 72.91 {\color{red}(-9.99)}
& 32.67 {\color{red}(-38.27)}
& 28.30 {\color{red}(-34.19)} \\

Gemma3~\cite{team2025gemma} & Gemma3-4B-IT
& 17.48 {\color{red}(-29.32)}
& 11.73 {\color{red}(-31.06)}
& 4.29 {\color{red}(-25.95)}
& 70.84 {\color{red}(-12.06)}
& 27.19 {\color{red}(-43.75)}
& 22.04 {\color{red}(-40.45)} \\

SNIFFER~\citep{qi2024sniffer} & Vicuna-13B
& \underline{37.90} {\color{red}(-8.90)}
& \underline{33.63} {\color{red}(-9.16)}
& -- 
& \underline{80.41} {\color{red}(-2.49)}
& \underline{67.49} {\color{red}(-3.45)}
& \underline{61.44} {\color{red}(-1.05)} \\

MMD-Agent~\citep{liu2024mmfakebench} & Qwen2.5-7B
& 34.78 {\color{red}(-12.02)}
& 29.48 {\color{red}(-13.31)}
& -- 
& -- 
& 62.78 {\color{red}(-8.16)}
& 52.83 {\color{red}(-9.66)} \\

\midrule
\textbf{BiMi (Ours)} & Gemma3-4B-IT
& \textbf{46.80} 
& \textbf{42.79}
& \textbf{30.24}
& \textbf{82.90}
& \textbf{70.94}
& \textbf{62.49} \\
\bottomrule
\end{tabular}

\vspace{-2mm}
\label{tab:combined_with_delta}
\end{table*}

We evaluate BiMi on BiMiBench and MMFakeBench~\citep{liu2024mmfakebench}.
On BiMiBench, the model performs \emph{six-class} misinformation classification, region-level tamper localization, and explanation generation.
For MMFakeBench, which does not provide subtitles, we follow its standard \emph{four-class} setting and supply minimal textual context via a unified prompt; subtitle-specific objectives are disabled while the rest of the pipeline remains unchanged.

\noindent
\textbf{Implementation Details.} We adopt Gemma-3~\citep{team2025gemma} as the base multimodal large language model and apply the proposed three-stage post-training pipeline: domain align, supervised fine-tuning (SFT) and GRPO-based reinforcement learning.
The vision encoder is kept frozen throughout post-training, while only the multimodal adapter and language-model parameters are updated.
All experiments are conducted on 48 GB NVIDIA V100 GPUs with a global batch size of 16 and an initial learning rate of $1\times10^{-5}$.
SFT is run for 3 epochs with early stopping based on validation loss, followed by 2 additional epochs of GRPO training.
We fix the random seed to 42 and use AdamW with a weight decay of $1\times10^{-2}$.
The final reward combines the format, classification, and localization components without introducing any weighting coefficients, i.e.,
$R_{\text{total}} = R_{\text{format}} + R_{\text{cls}} + R_{\text{loc}}$,
so that each objective contributes equally.



\noindent
\textbf{Baselines.} We compare BiMi against a range of strong multimodal baselines, including InternVL3~\citep{zhu2025internvl3exploringadvancedtraining}, Qwen3~\citep{qwen3}, LLaVA-1.6~\citep{liu2024llavanext}, and LLama3.1~\citep{grattafiori2024llama3herdmodels}, which represent leading approaches in vision-language modeling. We also include misinformation detection systems such as SNIFFER~\citep{qi2024sniffer} and MMD-Agent~\citep{liu2024mmfakebench}, designed for out-of-context and multimodal misinformation. All models are evaluated using the same inputs: the image with overlaid Chinese-English subtitles and a task-specific prompt. 

\noindent
\textbf{Evaluation Metrics.}
We report \emph{Accuracy} (ACC) and \emph{F1} for classification and \emph{IoU} for tampered-region localization. 
For explanation quality, we follow Liu et al.~\citep{liu2023llava} and compute \emph{BERTScore}~\citep{bert-score} between model outputs and pseudo-references generated by GPT-4o~\citep{hurst2024gpt}. 
Pseudo-references are produced via structured prompting with label-grounded templates and manually verified by five trained annotators to ensure factual correctness and strict alignment with the ground-truth manipulations. 
To confirm that BERTScore reflects human judgment, the same annotators also rated 400 randomly sampled explanations on a 5-point Likert scale across five ground-truth–aligned dimensions—\emph{subtitle alignment}, \emph{visual detail consistency}, \emph{reasoning consistency}, \emph{clarity and readability}, and \emph{completeness}. 
The high agreement between these human ratings (Fleiss' $\kappa = 0.71$) demonstrates the reliability and practical relevance of our evaluation protocol. More evaluation details are provided in the Appendix.

\subsection{Main Results}

\noindent
\textbf{Performance Comparison.} We evaluate BiMi across three key subtasks on BiMiBench—(1) misinformation detection, (2) tampered region localization, and (3) explanation generation—and additionally assess cross-dataset generalization on MMFakeBench. As summarized in Table~\ref{tab:combined_with_delta}, BiMi consistently outperforms all baseline MLLMs across every metric and benchmark. On BiMiBench, it achieves notable improvements over the strongest baseline, including a +8.9 gain in detection accuracy, a +15.9 improvement in localization IoU, and a +2.5 increase in BERTScore for explanation quality. These gains are further strengthened by our GRPO-based reward design, which enhances both localization precision and explanation specificity. BiMi also achieves the best performance on MMFakeBench (4-class classification task), demonstrating strong robustness and generalization beyond the training domain.

\noindent
\textbf{Computation and Efficiency.} End-to-end latency includes three stages: model forward pass, OCR subtitle extraction, and external retrieval. Pure model inference averages ~320 ms per sample, while enabling OCR and web search adds 0.8–1.2 s depending on network conditions—similar to typical retrieval-augmented MLLMs. To ensure robustness, BiMi adopts a graceful-degradation strategy: if OCR or retrieval times out, the system automatically falls back to internal reasoning with minimal performance impact.

\noindent
\textbf{Ablation Study.} We investigate the contributions of each stage in our pipeline using an ablation study.
(i)~\textit{Domain Alignment.} Removing this stage reduces accuracy by 2.35 and BERTScore by 4.48, indicating its role in grounding visual-textual reasoning on news content.
(ii)~\textit{Instruction Tuning.} Eliminating instruction tuning leads to a drastic performance drop, accuracy falls by 31.18 and BERTScore by 12.06, highlighting its importance in enabling bilingual understanding. 
(iii)~\textit{GRPO.} Without GRPO, the model fails to generate specific explanations tied to image or \begin{table}[htp]
\centering
\small
\caption{\small Table: Ablation results (\%). Red text shows a performance drop from the best variant. } 
\begin{tabular}{lcc}
\toprule
Design & Accuracy & BERTScore \\
\midrule
Full model & \textbf{48.60} & \textbf{82.90} \\
w/o Domain Align & 46.25 \textcolor{red}{(-2.35)} & 78.42 \textcolor{red}{(-4.48)} \\
w/o Instr. Tune & 17.42 \textcolor{red}{(-31.18)} & 70.84 \textcolor{red}{(-12.06)} \\
w/o GRPO & 42.90 \textcolor{red}{(-5.70)} & 74.81 \textcolor{red}{(-8.09)} \\
Answer First & 46.66 \textcolor{red}{(-1.94)} & 81.79 \textcolor{red}{(-1.11)} \\
\bottomrule
\end{tabular}
\vspace{-2mm}
\label{tab:ablation}

\end{table} subtitle content, reducing accuracy by 5.7 and BERTScore by 8.09.
(iv)~\textit{Output Order.} Reversing the output order (classify before explain) slightly lowers accuracy (–1.94) and BERTScore (–1.11), suggesting that generating explanations facilitates more reasoning.
Table~\ref{tab:ablation} summarizes the results. These findings confirm that all components are essential, with instruction tuning and GRPO being particularly critical for boosting accuracy and explanation quality.

\begin{figure*}[t]
\centering
\vspace{-2mm}
\includegraphics[width=0.99\linewidth]{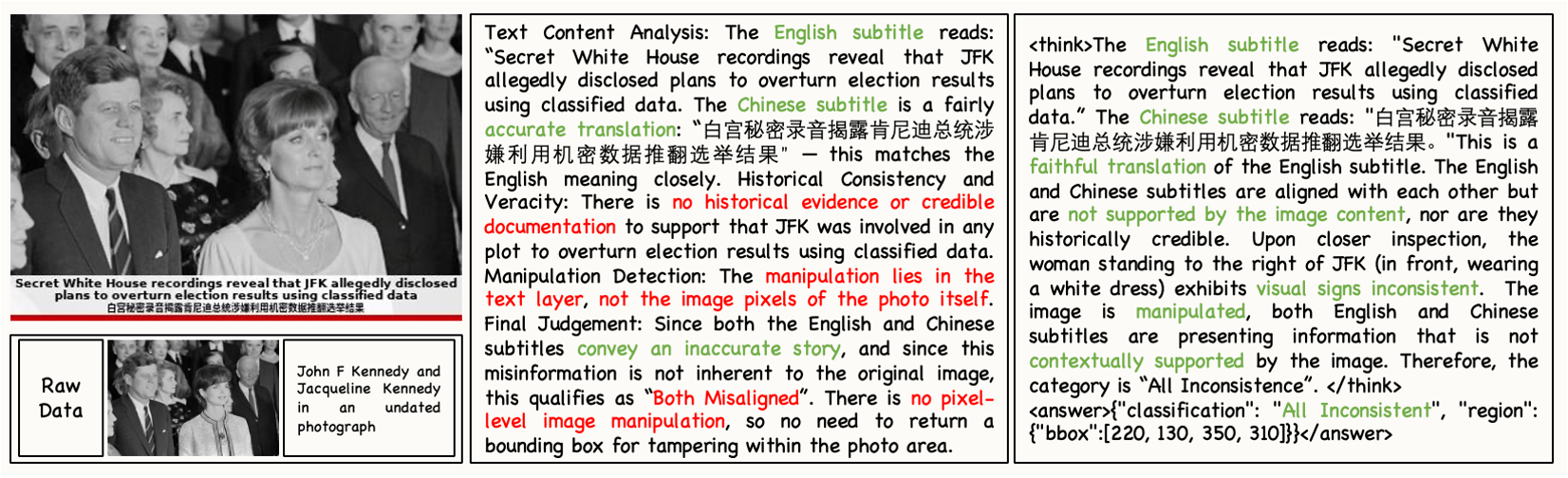}  

\caption{Comparison of explanations from InternVL3 (middle) and our model (right). Top left: input; bottom left: original sample. \textit{Some responses are truncated due to space constraints.}}

\vspace{-4mm}
\label{fig:example1}
\end{figure*}

\subsection{Explainability Analysis}

Beyond accurate classification and localization, generating high-quality natural language explanations is crucial for interpretable misinformation detection. We analyze explanation ability through qualitative comparisons, the influence of the retrieval module, and evaluation under bilingual.

\noindent
\textbf{Explanation Quality. }We compare BiMi with InternVL3~\citep{zhu2025internvl3exploringadvancedtraining} on a representative case (Figure~\ref{fig:example1}). Both models capture the bilingual subtitle alignment, but InternVL3 relies on external cues and misses the image–text mismatch. BiMi instead its decision in the image, correctly localizes the manipulation, and offers a cleare explanation.

\noindent
\textbf{Retrieval Module on Real Samples.}
We further tested the retrieval component on a 100-sample set of authentic bilingual news images to examine its behavior in real-world scenarios. Human evaluation shows that retrieved evidence improves explanation clarity or correctly resolves subtle cross-lingual inconsistencies in 9\% of cases, remains neutral in 89\%, and results in off-topic passages in only 2\%, typically due to OCR noise in Chinese subtitles. Importantly, even in these rare failure cases, the final predictions remain stable, indicating that retrieval errors do not propagate into the core reasoning pipeline. Overall, these results suggest that the retrieval module provides timely factual context that can meaningfully enrich the model’s explanations, while its downside risk is minimal—an encouraging property for practical deployment in noisy real-world environments.

\begin{table}[h]

\centering
\vspace{-1mm}
\caption{Accuracy and BERTScore for each language variant (4-category classification).}
\scalebox{0.9}{
\begin{tabular}{lcc}
\toprule
Input Variant & ACC & BERTScore \\
\midrule
CN-only           & 72.32 & -- \\
EN-only           & 82.48 & 83.44 \\
\bottomrule
\end{tabular}
}
\vspace{-3mm}
\label{tab:multilingual-inline}

\end{table}

\noindent
\textbf{Bilingual multimodal capability. }To assess bilingual understanding, we also create a single-language test variant where the model receives only Chinese or only English subtitles, reducing the task to a 4-class classification. Removing cross-lingual comparison makes the task substantially easier. As shown in Table~\ref{tab:multilingual-inline}, performance is noticeably higher with English-only inputs, reflecting the dominance of English data and reasoning patterns in pre-training. Chinese-only inputs introduce greater variance, particularly with idioms or long, multi-clause subtitles. These results reveal a clear gap in bilingual robustness and underscore the need for stronger Chinese modeling, improved tokenization, and more balanced multilingual data in future systems.





\noindent
\textbf{Failure Cases.}
BiMi’s remaining errors stem mainly from two sources. First, extremely subtle manipulations can lead to off-target localization or generic explanations, especially when visual edits or cross-lingual inconsistencies are barely perceptible. Second, OCR quality has a substantial impact: about 37\% of misclassifications result from truncated characters or rare-character errors in Chinese subtitles, issues rarely seen in English text. As shown in Table~\ref{tab:ocr_acc_bertscore}, performance drops noticeably when using OCR-extracted subtitles instead of ground-truth text, confirming OCR noise as a major contributor. More examples are in the Appendix.

\begin{table}[h]
\centering
\caption{Classification accuracy (ACC) and BERTScore (\%) 
using \emph{Ground-truth} subtitles (manually provided) versus 
\emph{OCR-extracted} subtitles (automatically recognized).}
\label{tab:ocr_acc_bertscore}
\begin{tabular}{lccc}
\toprule
Language & Subtitle Source & ACC (\%) & BERTScore \\
\midrule
English  & Ground-truth   & 84.24 & 88.10 \\
English  & OCR-extracted  & 83.02 & 85.92 \\
Chinese  & Ground-truth   & 80.07 & 86.54 \\
Chinese  & OCR-extracted  & 75.33 & 80.21 \\
\bottomrule
\end{tabular}
\vspace{-2mm}
\end{table}

\section{Conclusion and Limitation}

\textbf{Conclusion.} We study the task of detecting multimodal inconsistencies in news images with bilingual subtitles. To support this, we introduce BiMiBench, a large-scale benchmark with 104K samples featuring realistic visual edits and cross-lingual mismatches. We propose BiMi, a bilingual multimodal framework combining a three-stage post-training strategy and retrieval-augmented reasoning. Experiments show that BiMi achieves state-of-the-art performance across tasks, highlighting its effectiveness in real-world scenarios. This work advances research on bilingual and multimodal misinformation detection.

\noindent
\textbf{Limitation.} While BiMi achieves strong results on bilingual multimodal detection, several limitations remain.
First, it currently targets only Chinese–English subtitles; extending to additional languages would improve cross-lingual generalization.
Second, localization is restricted to bounding boxes and may miss fine-grained manipulations.
Third, the data diversity of BiMiBench is limited: beyond broader subtitle templates and editing styles, richer sources spanning multiple languages and topical domains are needed to better reflect real-world variability.
Fourth, retrieval introduces extra latency and depends on external sources, which can occasionally return noisy or outdated information.
Finally, BiMi’s reasoning capability could be strengthened, especially for cases with conflicting multimodal evidence.

\newpage
\appendix


\section*{Overview of Supplementary Material}






\noindent
\textbf{Section~\ref{sec:dataset}}: Dataset and ethics.

\noindent
\textbf{Section~\ref{sec:experiment}}: Additional experiments and cases.

\noindent
\textbf{Section~\ref{sec:train}}: Implementation details.
    
\section{Dataset Construction and Quality Control}
\label{sec:dataset}

\subsection{Dataset Statistics}

\begin{figure*}[h]
\centering
\includegraphics[width=0.99\linewidth]{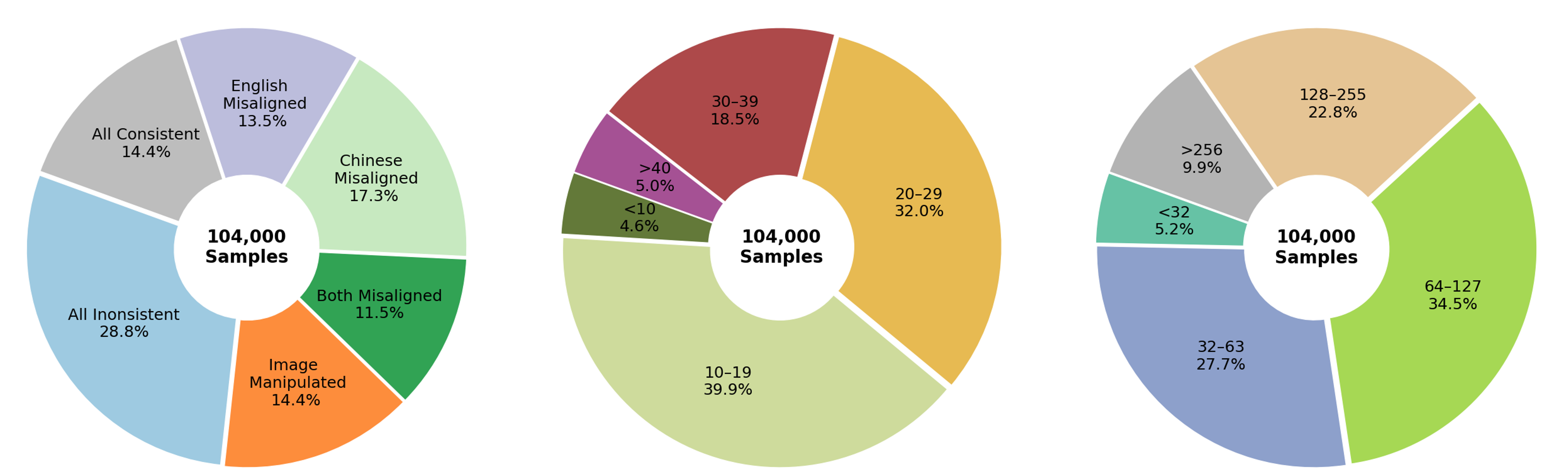}
\caption{Dataset statistics of BiMiBench. Left: distribution across six misinformation categories. Center: English subtitle length distribution. Right: normalized area of manipulated regions (bounding boxes height).}
\label{fig:dataset_stats}
\end{figure*}

BiMiBench comprises 104,000 samples spanning six fine-grained misinformation categories.
Figure \ref{fig:dataset_stats} summarizes the key dataset properties.
Beyond the balanced distribution across categories (left), BiMiBench exhibits substantial diversity in both visual content and bilingual subtitle structures.
The images cover a wide range of news domains, including politics, disasters, sports, daily life, technology, and international affairs, ensuring that manipulation patterns are not confined to a narrow set of scenes.
The subtitles vary in length, writing style, and translation quality across topics in both Chinese and English.
Such multimodal richness helps simulate real-world bilingual misinformation scenarios and provides a challenging, comprehensive testbed for fine-grained manipulation detection.

\subsection{Misinformation Generation Strategy.}
To simulate realistic multimodal misinformation, we generate manipulated subtitles by introducing subtle yet misleading semantic alterations. These edits preserve grammatical fluency and contextual plausibility while modifying key factual elements such as subjects, causal relations, locations, or numerical values. We further introduce bilingual inconsistencies by translating the manipulated English subtitles into Chinese, allowing semantic shifts to emerge naturally during translation. This strategy enables diverse and challenging misinformation scenarios that reflect real-world cross-modal and cross-lingual manipulations. We design task-specific prompts to guide misinformation generation, ensuring fluency and semantic plausibility while introducing factual inconsistencies; an example prompt used in this process is shown in Table~\ref{tab:misinfo_prompt}.

\begin{table}[h]\centering
\begin{minipage}{0.99\columnwidth}
\centering
\caption{Prompt templates used in BiMi for multimodal misinformation detection}
\begin{tcolorbox}[colback=white, colframe=black!50,
  boxsep=2pt, left=2pt, right=2pt, top=2pt, bottom=2pt] 
\textbf{$<$system$>$}

Based on the image, create a fictional, sensational  news headline or summary in one sentence. The story must be clearly fake but sound plausible, and must be no more than 30 words. The news should sound outrageous, scandalous, or unbelievable, yet still written in a serious, news-like tone.

\textbf{$<$/system$>$}
\end{tcolorbox}

\label{tab:misinfo_prompt}
\end{minipage}
\end{table}

\noindent
\textbf{Data Diversity.} To ensure broad and representative diversity in the synthetic misinformation, we introduce variation at three complementary stages. \textit{Data source}: we begin with the large and professionally curated VisualNews corpus, whose wide topical coverage—politics, economics, sports, culture—naturally provides a variety of writing styles and visual contexts, ensuring that manipulations are not limited to a narrow set of news domains. \textit{Editing}: on top of this diverse base we generate multiple kinds of factual distortions, ranging from single–word entity swaps to complex changes in temporal references, numerical values and causal relations. We also randomize prompt formulations and the sampling parameters of the generation model so that the manipulated bilingual subtitles exhibit heterogeneous and unpredictable inconsistencies. \textit{Label assignment}: after generation, each candidate sample is manually reviewed by trained annotators and assigned to one of the six defined misinformation categories, reflecting the precise combination of visual edits and cross-lingual subtitle inconsistencies. This three-stage process guarantees that BiMiBench covers a wide spectrum of realistic manipulation patterns and prevents overfitting to any type of false information.

\subsection{Review Protocol} 
To ensure data quality and annotation reliability, all samples in BiMiBench were manually verified by a team of five trained annotators, each with backgrounds in journalism, linguistics, or media studies, and prior experience with fact-checking or misinformation analysis. Annotators received targeted instruction on identifying realistic image manipulations, assessing semantic consistency across English-Chinese subtitle pairs, and applying fine-grained manipulation categories. During the review process, they checked the accuracy of tampered regions, validated bounding box alignment, and evaluated bilingual subtitles for cross-lingual fidelity and plausibility. Samples with OCR errors, ambiguous edits, mistranslations, or low visual quality were discarded. This multi-layered, expert-driven verification process ensures that BiMiBench maintains high annotation quality, semantic precision, and supports reliable benchmarking for multilingual multimodal misinformation detection.

\subsection{Ethical and Licensing Considerations}
BiMiBench is derived from the publicly available VisualNews corpus, which provides professionally curated image–text news pairs under a license permitting academic research.  All original images and captions remain the intellectual property of their respective news organizations; our release is strictly for non-commercial, research purposes and requires users to acknowledge the original sources and this benchmark in any derivative work.

While BiMiBench is designed to advance scientific understanding of multimodal misinformation detection, it necessarily contains intentionally manipulated content: images edited to introduce subtle visual changes and bilingual subtitles rewritten to create semantic inconsistencies.  This “dual nature’’—a resource built for research, yet embedding realistic examples of fabricated content—poses an inherent risk of misuse if taken out of context or circulated without clear academic framing.

To mitigate these risks, we (i) clearly mark all manipulated samples as synthetic, (ii) provide the data only for research and educational use, and (iii) require that any redistribution or downstream work include appropriate citations and comply with relevant copyright and data-protection regulations.  We explicitly prohibit using BiMiBench to create or disseminate deceptive media outside controlled research settings.  These safeguards aim to ensure that the benchmark remains a tool for studying and countering misinformation rather than a source of it.
\subsection{Example Visualization.}
\begin{figure*}[]
\centering
\includegraphics[width=0.99\linewidth]{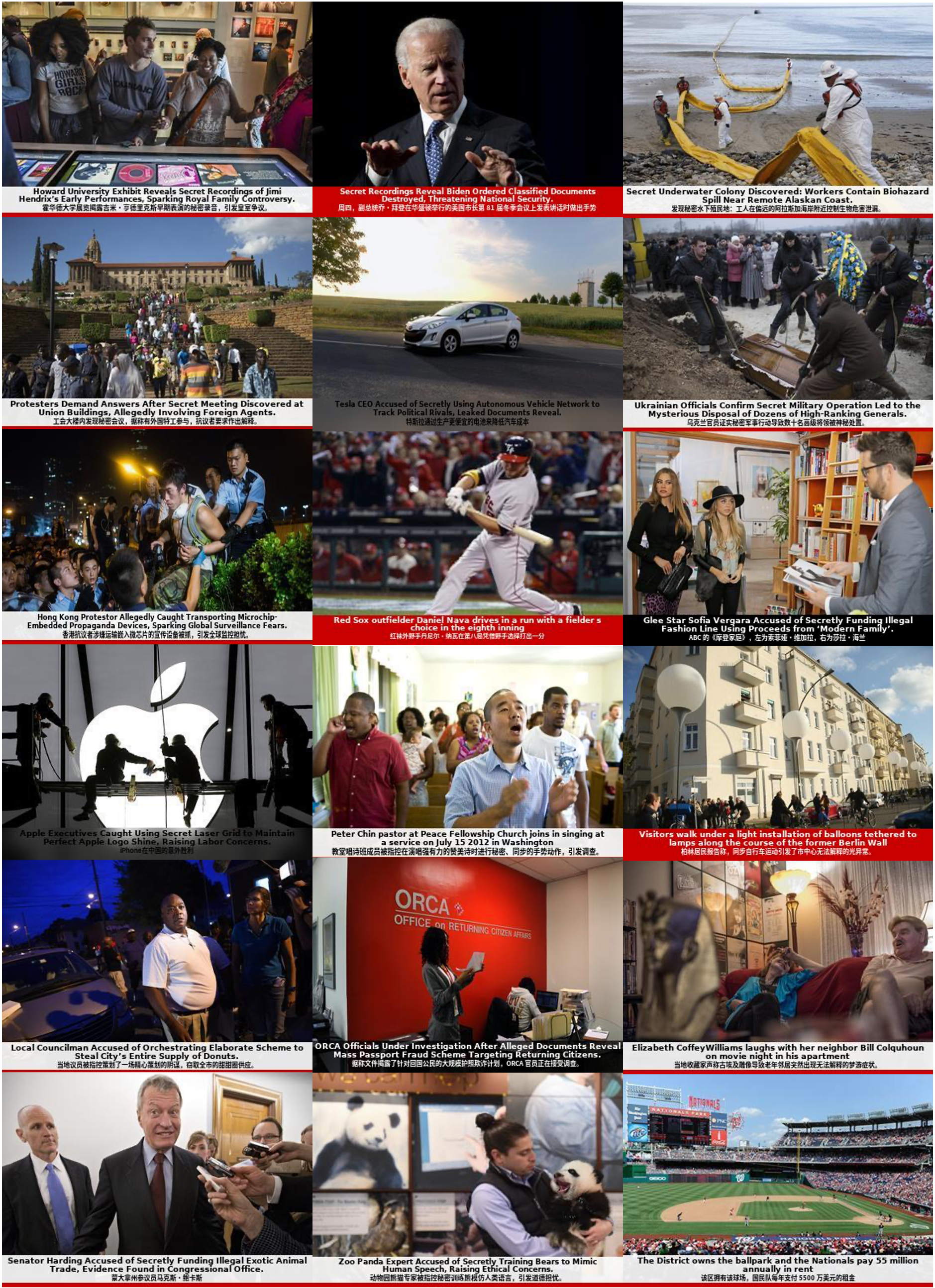}
\caption{Sample BiMiBench examples with bilingual subtitles and visual or text manipulations.}
\label{fig:example}
\end{figure*}

Figure~\ref{fig:example} illustrates representative BiMiBench samples, showcasing a wide range of manipulation patterns, cross-lingual inconsistencies, and subtitle complexities.

\section{Additional Experiments}
\label{sec:experiment}
\subsection{Reference Generation}
To construct high-quality reference explanations for automatic evaluation, we adopt a two-stage human–LLM pipeline.
First, for each sample we prompt GPT-4o with task-specific, label-grounded templates to generate \emph{two} candidate explanations conditioned on the manipulated modality and misinformation type.
Second, a team of five trained annotators—each with a background in journalism, media studies, or NLP—independently reviewed the two candidates and selected the one that best satisfied five ground-truth–aligned dimensions:
(i) \emph{Subtitle alignment}, (ii) \emph{Visual detail consistency}, (iii) \emph{Reasoning consistency}, (iv) \emph{Clarity and readability}, and (v) \emph{Completeness}.
The chosen explanation for each sample was retained as the pseudo-ground truth used in BERTScore evaluation.
This procedure ensures that every reference explanation is both fluent and factually faithful while keeping the generation process efficient and reproducible.

\begin{table*}[t]
\centering
\caption{Human evaluation criteria and scoring scheme for selecting reference explanations. Annotators scored each criterion on a 0–2 scale
(0 = not satisfied, 1 = partially satisfied, 2 = fully satisfied),
and explanations with a total score below 7 were discarded or revised.}
\label{tab:criteria}
\small
\setlength{\tabcolsep}{6pt} 
\begin{tabular}{p{3.5cm} p{11.0cm} c}
\toprule
\textbf{Criterion} & \textbf{Description} & \textbf{Score} \\
\midrule
Subtitle alignment 
& Checks consistency between English–Chinese subtitles and the described manipulation.
& 0--2 \\

Visual detail consistency
& Ensures the explanation correctly reflects visible tampered regions or edits.
& 0--2 \\

Reasoning consistency
& Evaluates logical coherence between detected evidence and the explanation.
& 0--2 \\

Clarity and readability
& Measures fluency and conciseness suitable for news-style explanations.
& 0--2 \\

Completeness
& Ensures all manipulated elements are covered without omitting key evidence.
& 0--2 \\
\bottomrule
\end{tabular}
\end{table*}

\subsection{Performance on Real-World Samples}
To assess real-world generalization, we manually collected 100 recent social-media posts spanning political, health, and international news.  
These posts contain image–caption manipulations, cross-lingual inconsistencies, and subtle visual edits typical of in-the-wild misinformation.  
Because the authenticity of such posts cannot be verified with ground-truth labels, the 100 examples were \emph{expert-curated}: five annotators with backgrounds in journalism and fact-checking jointly reviewed the content and reached consensus on whether each case constitutes misinformation.  
This human consensus served as the reference label set.

Evaluated against these expert judgments, BiMi correctly identified 81 of the 100 curated cases (81\% accuracy) .  
These findings suggest that BiMi can transfer effectively from synthetic benchmarks to uncontrolled real-world conditions, even when the ground truth is based on expert assessment rather than definitive factual verification.

\begin{figure*}[h]
\centering
\includegraphics[width=\textwidth]{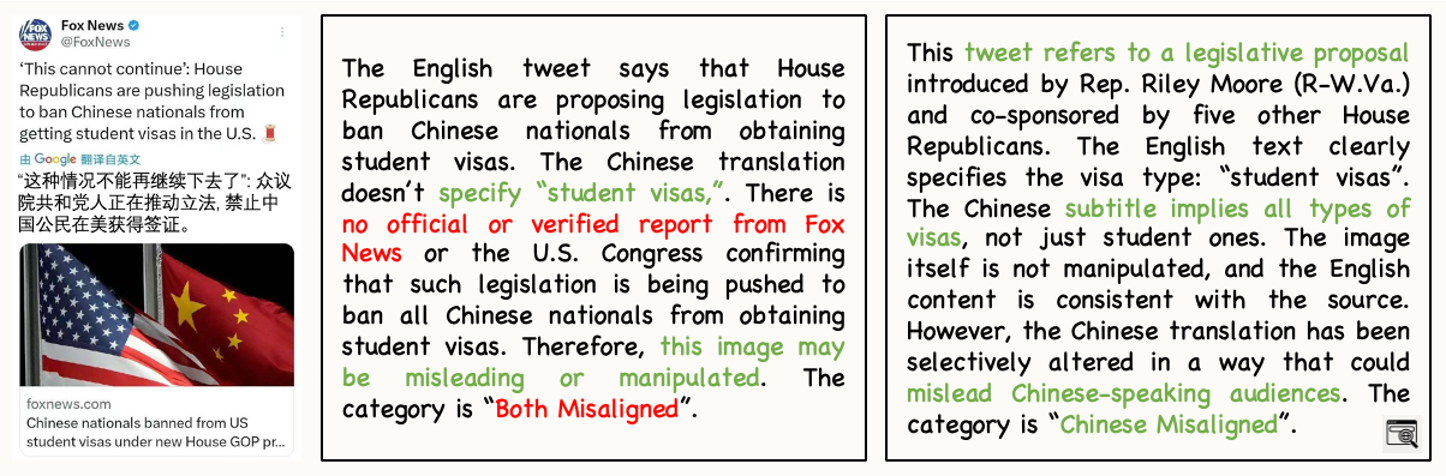} 
\caption{Social media example with subtle bilingual inconsistency. The Chinese translation omits "student", changing the meaning and potentially misleading Chinese readers. Middle: InternVL3 result; Right: BiMi result. }
\label{fig:qualitative}
\end{figure*}

We analyze a real-world social media post with subtle bilingual inconsistency (Figure~\ref{fig:qualitative}). The removed "student" in the Chinese translation alters the meaning. Without retrieval, the model misclassifies due to a lack of context. The retrieval result provides a key clue for classification. BiMi correctly identifies the inconsistency, showing its benefit for nuanced cross-lingual reasoning.

\subsection{Effect of OCR Module}
This section provides brief supplementary observations expanding on the OCR analysis in the main paper. While the core quantitative results are already reported, here we include additional qualitative notes.

During manual inspection, we found that OCR errors most commonly occur in
Chinese subtitles containing long sentence structures or low-resolution text, which aligns with the degradation patterns shown in the main paper. In contrast, English subtitles are generally recognized cleanly, leading to smaller performance differences between ground-truth and OCR-extracted inputs. In many real-world samples, OCR inaccuracies do not alter the final classification category but may affect explanation specificity. We therefore recommend simple preprocessing steps—such as cropping subtitle regions or modest contrast enhancement—to stabilize OCR quality when applying the model outside the benchmark setting.

These findings show that BiMi remains largely robust to moderate OCR noise; the observed degradation originates mainly from preprocessing rather than from limitations in the model’s bilingual reasoning.  When clean Chinese subtitles are provided, BiMi’s accuracy approaches the English benchmark. Improving OCR fidelity can enhance overall system reliability in real-world deployments.

\subsection{Analysis of Bilingual Performance Gap}
BiMi achieves higher accuracy on English-only inputs (82.48\% ACC) than on Chinese-only inputs (72.32\% ACC).
To disentangle the causes of this gap, we conducted a controlled evaluation using ground-truth subtitles versus OCR-extracted subtitles, as summarized in Table~\ref{tab:ocr_impact_1}.

\begin{table}[h]
\centering
\caption{Impact of OCR on classification accuracy.}
\label{tab:ocr_impact_1}
\setlength{\tabcolsep}{3pt}  
\begin{tabular}{lccc}
\toprule
Language & GT (\%) & OCR (\%) & Drop (\%) \\
\midrule
English & 84.2 & 83.0 & \textcolor{red}{-1.2} \\
Chinese & 80.0 & 75.3 & \textcolor{red}{-4.7} \\
\bottomrule
\end{tabular}
\end{table}

The results show that Chinese performance is far more sensitive to OCR noise, with a 4.7\% drop compared to only 1.2\% for English.
Thus, OCR quality in Chinese subtitles is the primary factor behind the overall 10\% gap.

Secondary contributors include (i) training data imbalance: VisualNews contains more English captions, which biases the backbone model toward English, and (ii) language-model priors: although Gemma 3 supports bilingual reasoning, its representations are stronger for English.
When clean Chinese subtitles are provided, BiMi’s accuracy approaches the English benchmark, confirming that OCR errors are the dominant bottleneck.

To close this gap, future work will focus on (1) enhancing Chinese OCR through confidence-based filtering and multi-OCR consensus, (2) exploring LLM-based subtitle repair to recover noisy outputs, and (3) fine-tuning on more balanced bilingual data.
Overall, the performance difference is mainly driven by OCR noise, with data imbalance and model priors as secondary factors.

\subsection{Computation and Efficiency}
BiMi employs a lightweight multimodal backbone (Gemma 3–4B) to maintain manageable training and inference costs. All experiments are conducted on a single NVIDIA V100 48 GB GPU. End-to-end inference latency is determined by three components: the model forward pass, OCR-based subtitle extraction, and the external retrieval module. Without retrieval, pure model inference averages 320 ms per sample. Enabling the full retrieval pipeline—including OCR and Google Search API queries—adds approximately 0.8–1.2 s of overhead, depending on network conditions and hardware. This additional cost remains acceptable for offline evaluation and is broadly consistent with other retrieval-augmented MLLMs.

To avoid retrieval becoming a single point of failure, BiMi incorporates a graceful-degradation mechanism: if OCR or web search times out or returns no relevant context, the model skips the retrieved evidence and falls back to pure visual–textual reasoning, keeping classification and explanation performance stable.
During training, the vision encoder is frozen and only LoRA adapters are updated, which keeps GPU memory usage below 40 GB and allows a batch size of 16.
These design choices ensure that BiMi can be reproduced on standard academic hardware while maintaining reasonable inference latency and memory footprint.

\subsection{Failure Case Study}
\label{sec:failure}
Despite the overall effectiveness of our model, we observed several notable failure cases that highlight its current limitations. One recurring issue involves semantic misinterpretation, particularly when the input contains rare or ambiguous phrases. For example, in a prompt requiring multi-step reasoning or contextual understanding (e.g., “Describe the implication of climate change on coastal microeconomies”), the model occasionally generates generic or overly broad responses that fail to capture the specific nuances of the query. Another type of failure arises in edge cases involving domain-specific knowledge. 

\begin{figure*}[h]
\centering
\includegraphics[width=0.99\textwidth]{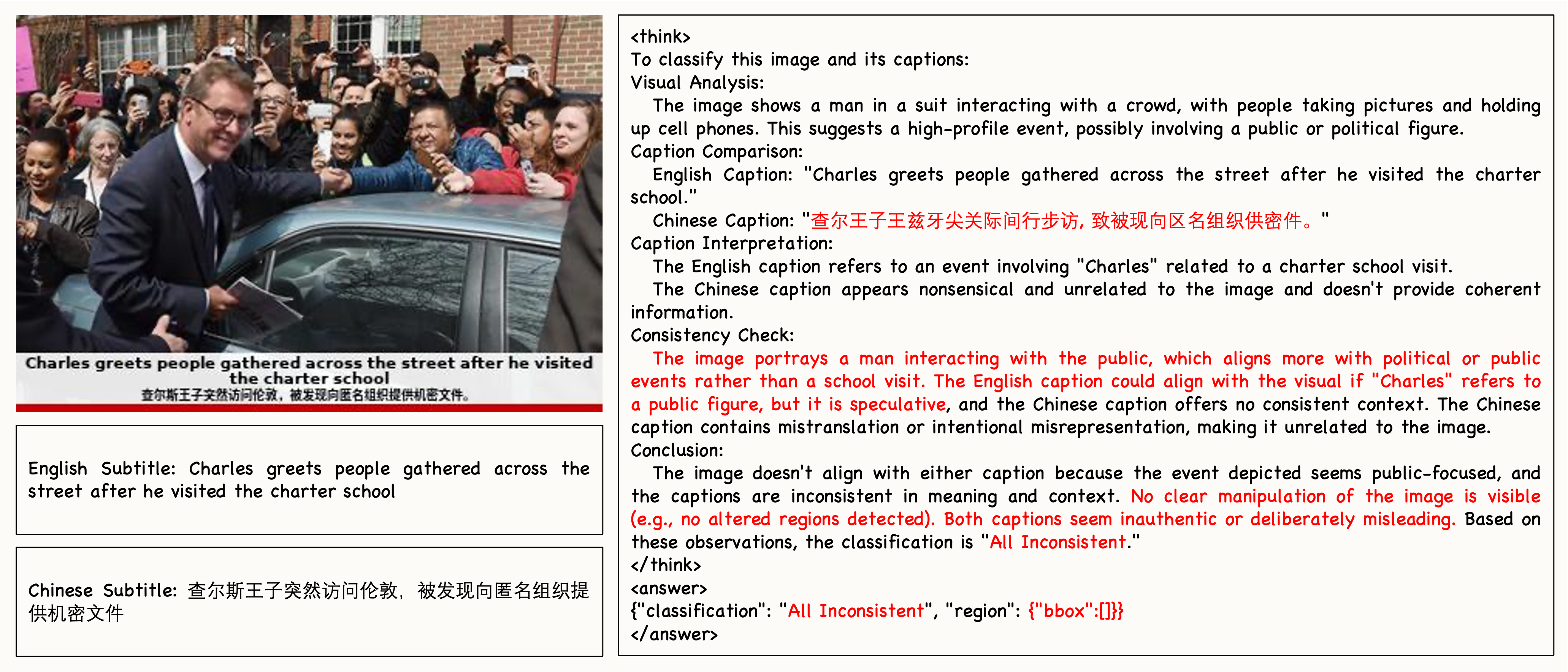}
\caption{Failure case}
\label{fig:failure}
\end{figure*}

Figure~\ref{fig:failure} shows a failure case where the model misclassified an image manipulation example as “All Inconsistent.” The English caption aligns with the image, but the Chinese one introduces a false espionage narrative. Key issues include (1) misclassification, (2) incorrect caption judgment (only the Chinese caption was wrong), and (3) a missing bounding box.

\subsection{Qualitative Cases}
\begin{figure*}[h]
\centering
\includegraphics[width=0.99\textwidth]{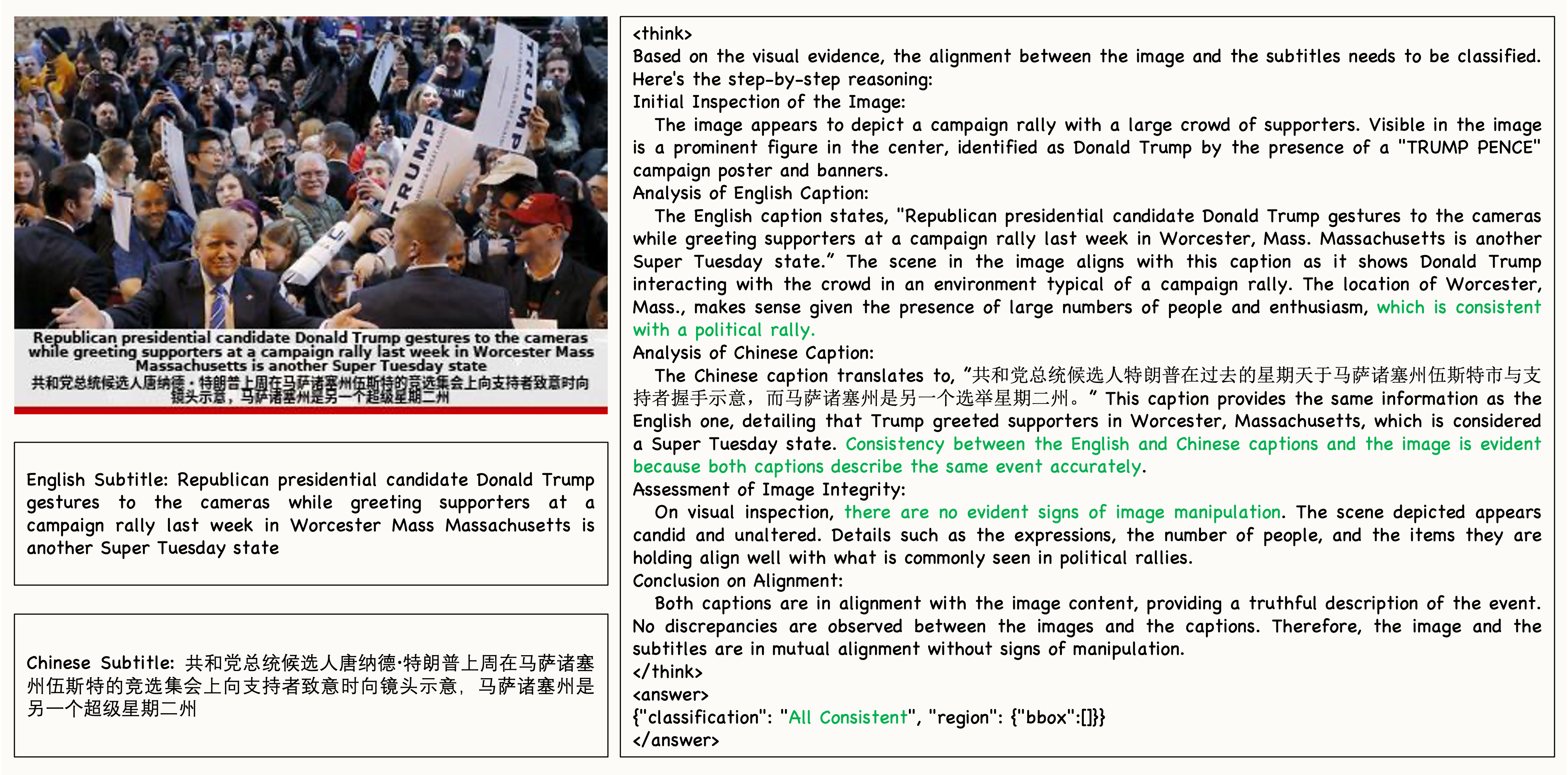}
\caption{Case of All Consistent.}
\label{fig:e1}
\vspace{-1.5em}
\end{figure*}

\begin{figure*}[h]
\centering
\includegraphics[width=0.99\textwidth]{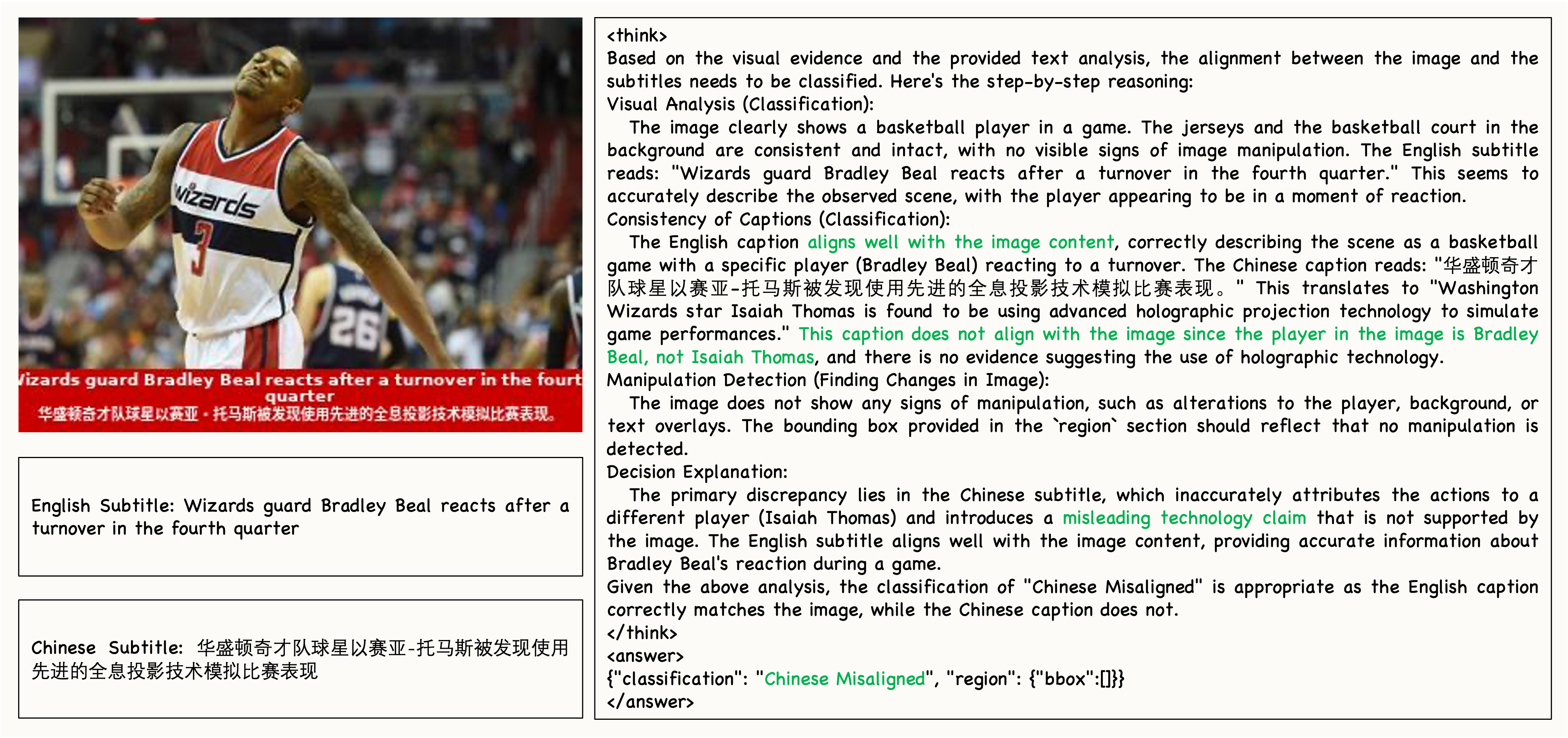}
\caption{Case of Chinese Misaligned.}
\label{fig:e2}
\vspace{-1.5em}
\end{figure*}

\begin{figure*}[h]
\centering
\includegraphics[width=0.99\textwidth]{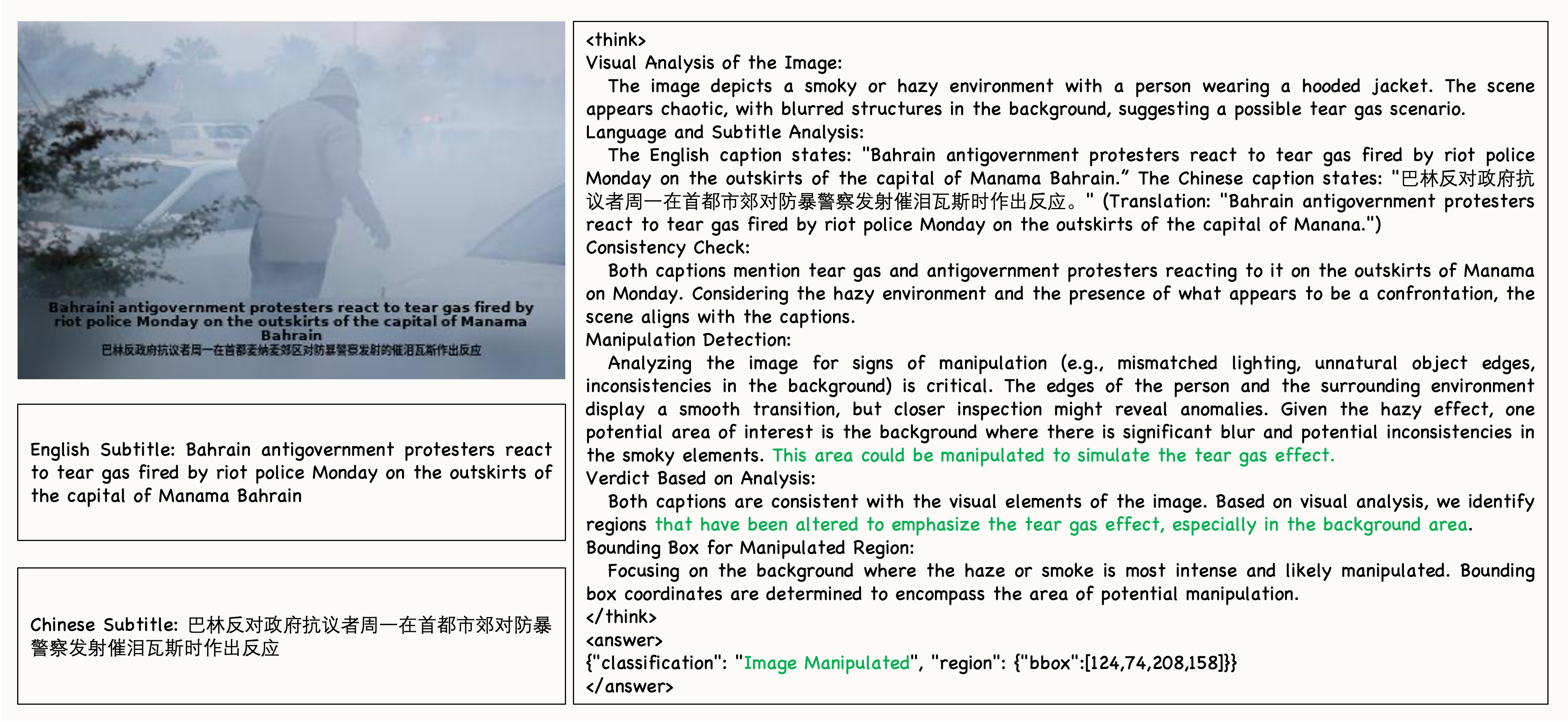}
\caption{Case of Image Manipulated.}
\label{fig:e3}
\vspace{-1.5em}
\end{figure*}

\begin{figure*}[h]
\centering
\includegraphics[width=0.99\textwidth]{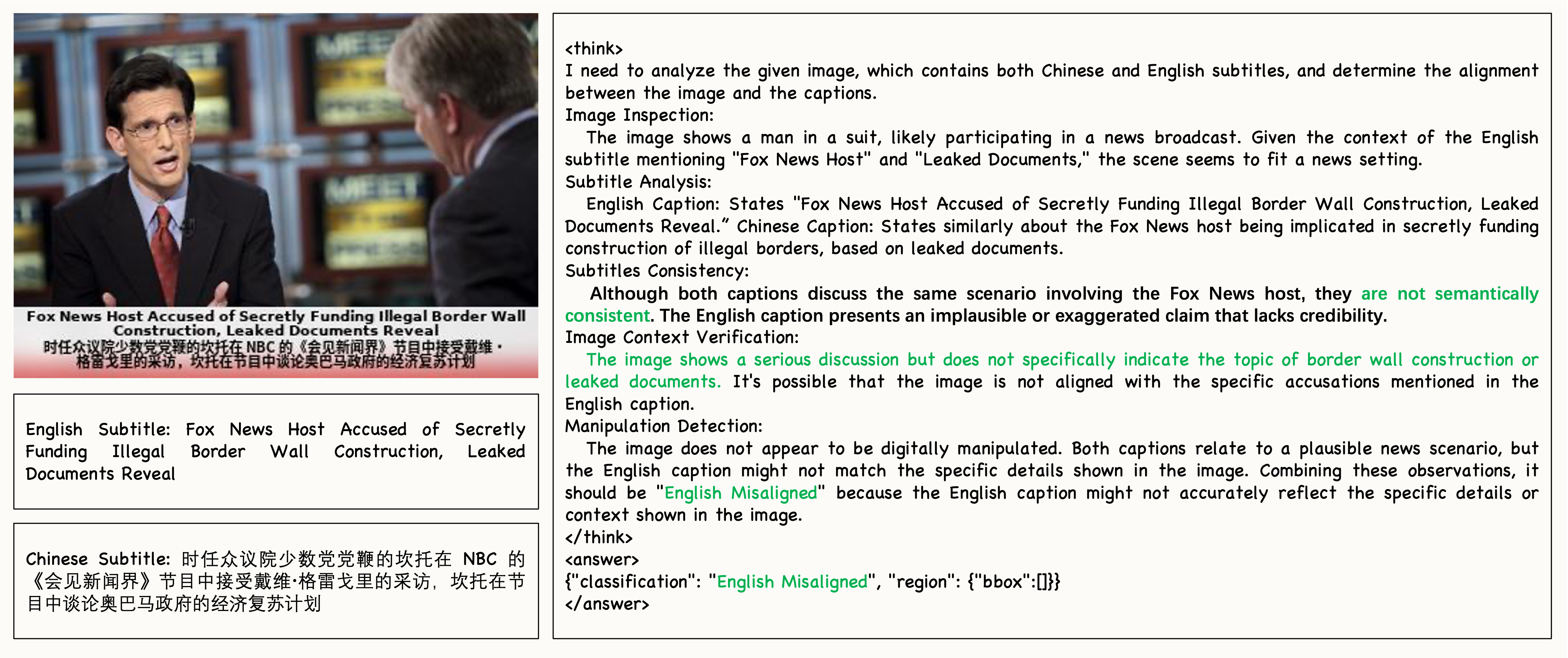}
\caption{Case of English Misaligned.}
\label{fig:e4}
\end{figure*}

\begin{figure*}[h]
\centering
\includegraphics[width=0.99\textwidth]{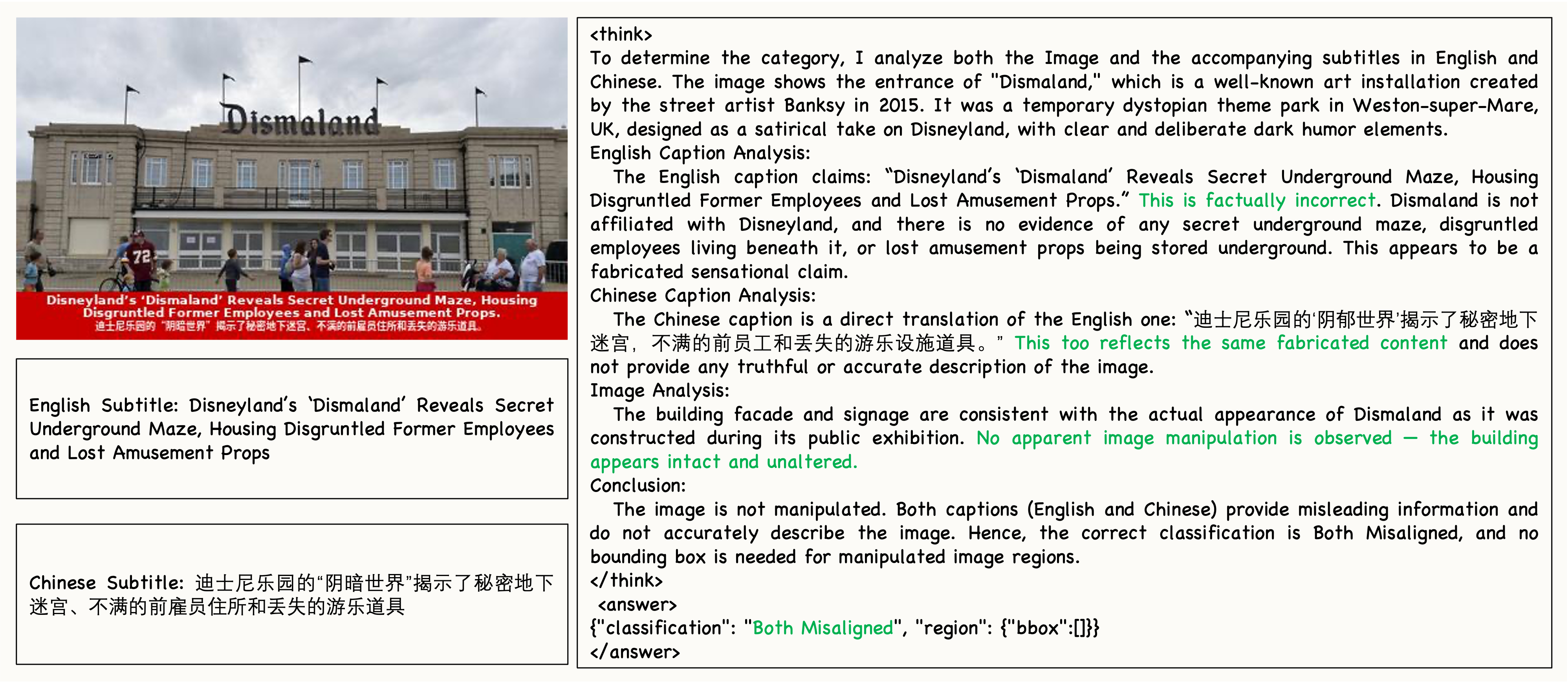}
\caption{Case of Both Misaligned.}
\label{fig:e5}
\end{figure*}

\begin{figure*}[h]
\centering
\includegraphics[width=0.99\textwidth]{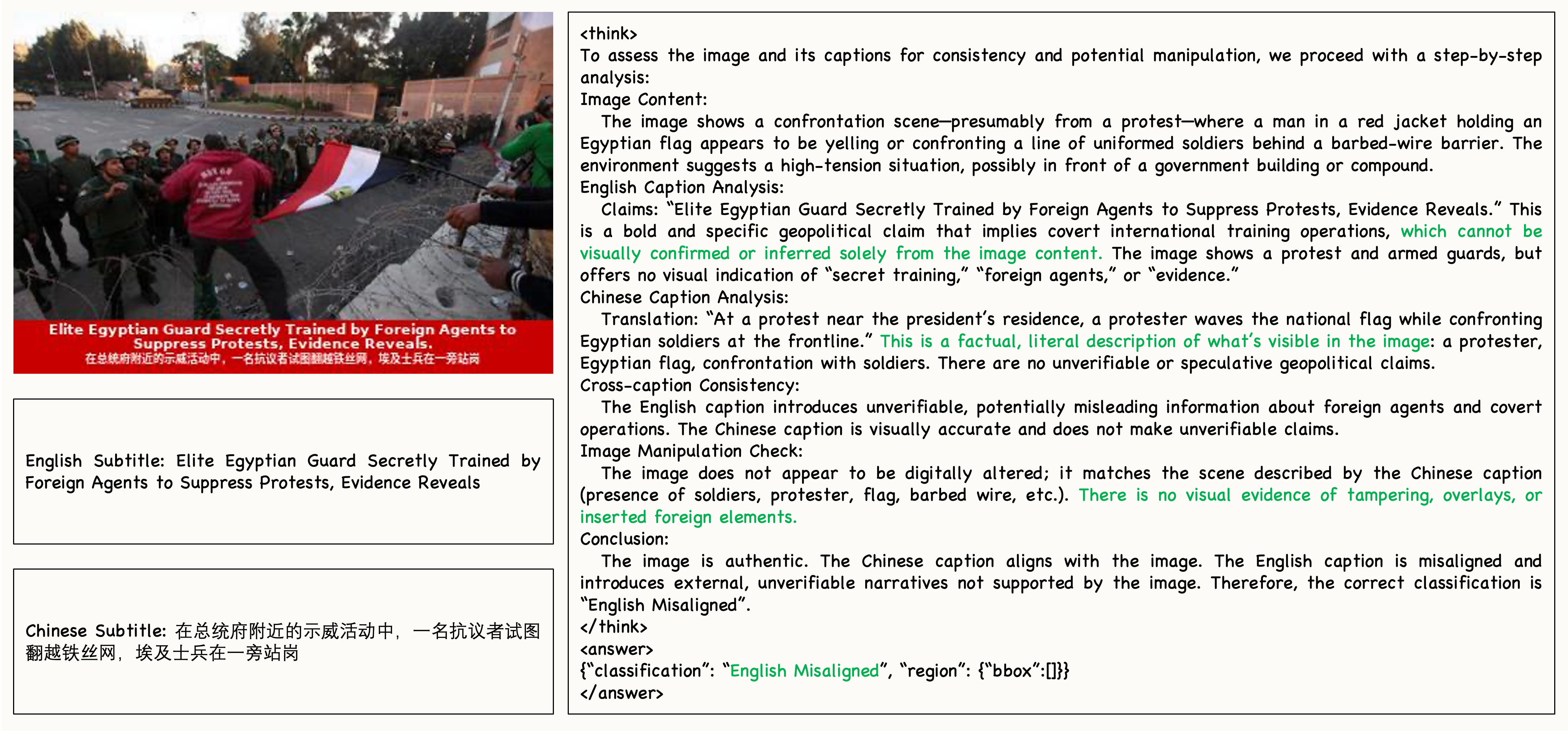}
\caption{Case of English Misaligned.}
\label{fig:e6}
\end{figure*}

\begin{figure*}[h]
\centering
\includegraphics[width=0.99\textwidth]{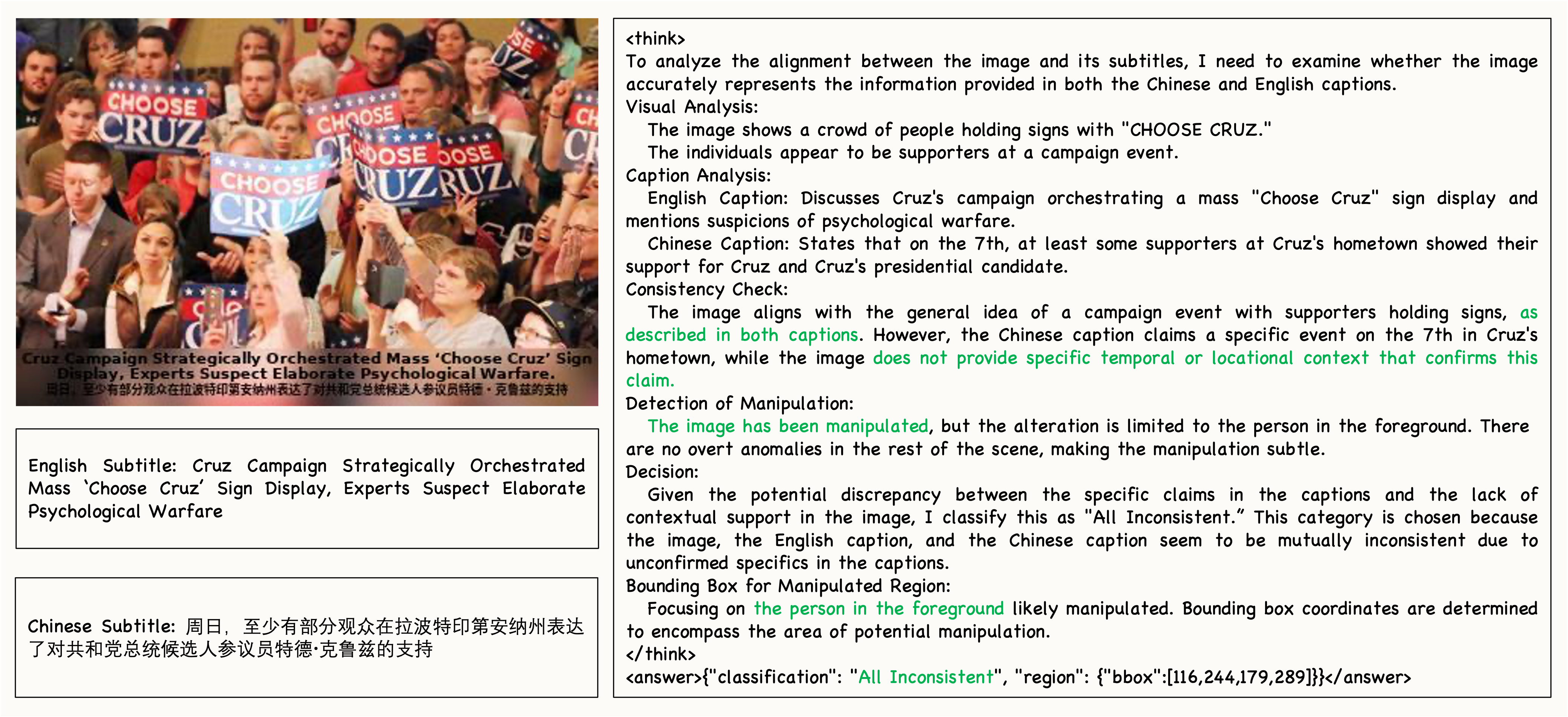}
\caption{Case of All Inconsistent.}
\label{fig:e7}
\end{figure*}

\begin{figure*}[h]
\centering
\includegraphics[width=0.99\textwidth]{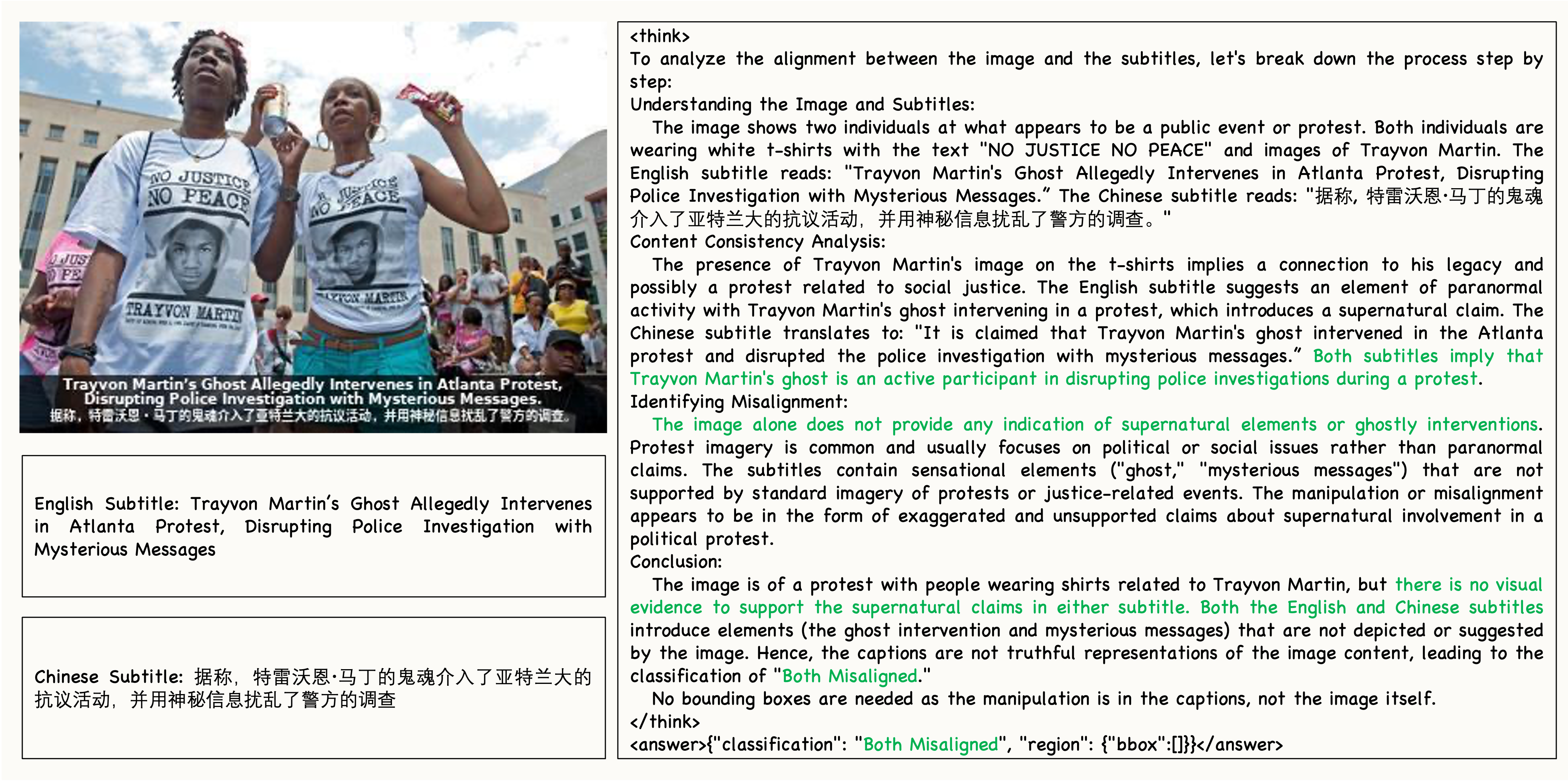}
\caption{Case of Both Misaligned.}
\label{fig:e8}
\end{figure*}

Figures~\ref{fig:e1}, \ref{fig:e2}, \ref{fig:e3}, \ref{fig:e4},
\ref{fig:e5}, \ref{fig:e6}, \ref{fig:e7}, \ref{fig:e8} show the examples.
\label{sec:example}

\section{Implementation Details}
\label{sec:train}

\subsection{Training}
\textbf{Model Architecture.}
Our framework is built upon the publicly released \textbf{Gemma 3} multimodal language model.
The vision encoder is initialized from a pretrained ViT-L and kept frozen in all training stages, while LoRA adapters are fine-tuned on top of the language backbone to reduce computational cost.
The model accepts as input the news image with its English–Chinese subtitles, serialized into a unified prompt that embeds the subtitles as text tokens and the image as patch-level embeddings.
Visual tokens from the frozen encoder are projected into the language model’s embedding space and fused with textual tokens through the cross-modal attention layers of Gemma 3, enabling joint reasoning over visual regions and bilingual text.
This design allows efficient adaptation to our task while preserving the strong multilingual and vision–language alignment of the pretrained model.

\textbf{Tokenization and Input Format.}
Both English and Chinese subtitles are tokenized using the SentencePiece tokenizer aligned with the Gemma 3 backbone to ensure consistent multilingual encoding.
The input is formatted into a structured prompt that specifies the prediction task and clearly marks different modalities; visual features from the frozen ViT-L encoder are projected into the same embedding space and appended as a sequence of visual tokens.
The combined text and image tokens are then fed to the cross-modal transformer, with the total input length (text + visual tokens) capped at 512 text tokens plus the visual tokens.

\textbf{Hyperparameter Summary.}
We fine-tune BiMi using the Gemma 3 model (4B) with a batch size of 8, a learning rate of \(1 \times 10^{-5}\), and a maximum input length of 512 tokens. The model is optimized with AdamW, using \(\beta_1 = 0.9\), \(\beta_2 = 0.98\), and a weight decay of 0.01. To stabilize training, we use gradient accumulation (4 steps) and mixed precision (FP16). The GRPO reward coefficients \(\lambda\)s are all set to 1. These hyperparameters are kept consistent across all three training stages unless otherwise specified.

\textbf{Prompt Template.} We design a structured instruction format to guide the model in reasoning across modalities and generating faithful explanations. The prompt template used during instruction tuning and inference is shown in Table~\ref{tab:prompt}. This format ensures consistent output structure, supporting multimodal reasoning (via \texttt{<think>}), interpretable classification, and region-level manipulation localization.

\subsection{Retrieve Module}
To improve adaptability to rapidly emerging misinformation, we design a retrieval module that operates \emph{only} at inference time and does not require additional training.
First, we apply OCR to extract both the Chinese and English subtitles embedded in the news image.
The two subtitles are concatenated to form a bilingual query, which is then sent to the Google Search API.
The API returns up to the top-3 relevant passages; if no relevant documents are found, the retrieval component simply returns an empty string.
The retrieved text snippets are concatenated and prepended to the model’s input prompt, yielding a unified representation $\mathcal{P}=\operatorname{concat}(R,I,S)$, where $R$, $I$ and $S$ denote the retrieved context, the image and the bilingual subtitles, respectively.
This augmented prompt is fed to the MLLM backbone so that cross-attention layers can jointly encode the external evidence and the visual–textual cues.
Because the module only provides auxiliary context and does not participate in training, the final predictions remain primarily grounded in the image–subtitle content while benefiting from timely external information when available.

\clearpage

\begin{table*}[h]\centering
\begin{tabular}{p{0.95\textwidth}}
\caption{Prompt templates used in BiMi for multimodal misinformation detection}
\centering
\begin{tcolorbox}[colback=white, colframe=black!50,
  boxsep=2pt, left=2pt, right=2pt, top=2pt, bottom=2pt] 
\textbf{$<$system$>$}

You are an expert in misinformation detection area. A conversation between User and Assistant. The user asks a question, and the Assistant solves it. The assistant first thinks about the reasoning process in the mind and then provides the user with the answer. The reasoning process and answer are enclosed within $<$think$>$ $<$/think$>$ and $<$answer$>$ $<$/answer$>$ tags, respectively, i.e., $<$think$>$ reasoning process here $<$/think$>$$<$answer$>$ answer here $<$/answer$>$. 

Assign one of the following six categories based on their mutual alignment: All Consistent: The image aligns well with both Chinese and English captions. No signs of manipulation or misalignment. Image Manipulated: The image is manipulated. Both captions truthfully describe the manipulated image. Both Misaligned: Both Chinese and English captions are manipulated. Neither caption correctly describes the image. Chinese Misaligned: Only the Chinese caption is manipulated. The English caption aligns correctly with the image. English Misaligned: Only the English caption is manipulated. The Chinese caption aligns correctly with the image. All Inconsistent: The image, Chinese caption, and English caption are all manipulated and mutually inconsistent. 

\textbf{$<$/system$>$}

\textbf{$<$user$>$}

Please analyze the given image containing both Chinese and English subtitles and complete the following three tasks:

(1) Classification Task: classify the alignment between the image and the subtitles into one of the following six categories: \texttt{"all consistent"}, \texttt{"image manipulated"}, \texttt{"both subtitles misaligned with image"}, \texttt{"only English aligned"}, \texttt{"only Chinese aligned"}, \texttt{"all inconsistent"}.

(2) Manipulation Detection: if the image has been manipulated, return one or more bounding boxes for the manipulated regions in the format: \texttt{\{"bbox":[x\_min, y\_min, x\_max, y\_max]\}}. If no manipulation is found, return an empty list.

(3) Decision Explanation: briefly explain your thinking before the classification and any detected regions.

Return your output using the following format, wrapped in tags:

$<$think$>$

Your explanation here.

$<$/think$>$

$<$answer$>$

\texttt{{"classification": "result", "region": [\{"bbox": [x\_min, y\_min, x\_max, y\_max]\}]}}

$<$/answer$>$

\textbf{$<$/user$>$}
\end{tcolorbox}

\label{tab:prompt}
\end{tabular}
\end{table*}

\clearpage

{
    \small
    \bibliographystyle{ieeenat_fullname}
    \bibliography{main}
}


\end{document}